\documentclass[runningheads]{llncs}

\usepackage[mobile]{eccv}

\usepackage{eccvabbrv}

\usepackage{graphicx}
\usepackage{tabularx} 
\usepackage{booktabs}
\usepackage[table]{xcolor} 
\usepackage{adjustbox} 
\usepackage{enumitem}

\usepackage[accsupp]{axessibility}  

\usepackage{hyperref}

\usepackage{orcidlink}

\usepackage{microtype}
\usepackage{graphicx}
\usepackage{subcaption}
\usepackage{booktabs} 
\usepackage{multirow}
\usepackage{comment}
\usepackage{tcolorbox}
\tcbuselibrary{breakable}
\usepackage{sidecap}
\sidecaptionvpos{figure}{t}

\usepackage{hyperref}
\usepackage{float}

\usepackage{amsmath}
\usepackage{amssymb}
\usepackage{mathtools}

\usepackage{amsthm}
\usepackage{comment}

\usepackage{xcolor}
\usepackage{booktabs}
\usepackage{colortbl}

\definecolor{posgain}{HTML}{008000} 
\definecolor{negloss}{HTML}{CC0000} 
\usepackage{booktabs}
\usepackage{colortbl}
\usepackage{xcolor}
\usepackage{threeparttable}

\definecolor{highlight}{gray}{0.94}
\definecolor{best}{gray}{0.85}
\definecolor{runnerup}{gray}{0.92}
\definecolor{groupgray}{gray}{0.97}

\usepackage[capitalize,noabbrev]{cleveref}
\usepackage[misc]{ifsym} 

\theoremstyle{plain}

\theoremstyle{definition}

\theoremstyle{remark}
\usepackage[textsize=tiny]{todonotes}

\begin{document}

\title{LaVPR: Benchmarking Language and Vision for Place Recognition} 

\titlerunning{LaVPR: Benchmarking Language and Vision for Place Recognition}
\authorrunning{O. Idan et al.}


\author{Ofer Idan \and Dan Badur \and Yosi Keller \and Yoli Shavit\thanks{Corresponding Author.}}
\institute{Faculty of Engineering, Bar-Ilan University \\ 
Ramat Gan, Israel \\
\email{\{ofer.idan, badur.dan, yosi.keller, yoli.shavit\}@biu.ac.il}}

\maketitle

\begin{abstract}
Visual Place Recognition (VPR) often fails under extreme environmental changes and perceptual aliasing. Beyond these limitations, standard systems cannot perform 'blind' localization from verbal descriptions alone, a capability critical for applications such as emergency response. To address these challenges, we introduce LaVPR, a large-scale benchmark that extends existing VPR datasets with over 650,000 rich natural-language descriptions. Using LaVPR, we investigate two paradigms: Multi-Modal Fusion for enhanced robustness and Cross-Modal Retrieval for language-based localization. Our results show that language descriptions yield consistent gains in visually degraded conditions, with the most significant impact on smaller backbones. Notably, adding language allows compact models to rival the performance of much larger vision-only architectures. For cross-modal retrieval, we establish a baseline using Low-Rank Adaptation (LoRA) and Multi-Similarity loss, which substantially outperforms standard contrastive methods across vision-language models. Ultimately, LaVPR enables a new class of localization systems that are both resilient to real-world stochasticity and practical for resource-constrained deployment.  Our dataset and code are available at \url{https://github.com/oferidan1/LaVPR}. 

\end{abstract}

\section{Introduction}\label{sec:introduction}
Traditional Visual Place Recognition (VPR) relies on matching global image descriptors to retrieve geo-tagged database entries. However, even state-of-the-art visual models struggle under extreme environmental variations, such as adverse weather, long-term temporal shifts, and motion blur~\cite{lu2024cricavpr}. To achieve robust, interactive localization, we propose extending the VPR framework with natural language, focusing on two key paradigms:
\begin{enumerate}[label=\roman*)]
    \item \textit{Multi-Modal Fusion ($V+L$)}: This approach utilizes language as a stable, high-level descriptor to disambiguate visual scenes. By incorporating semantic scene elements and visible text directly into a descriptive narrative, such as \textit{"a brick building featuring a 'Pharmacy' sign above the entrance"}, the system gains a viewpoint-invariant context. While visual degradation can compromise pixel-level features, these linguistic anchors provide a stable reference that maintains retrieval accuracy where vision-only models fail.
    \item \textit{Cross-Modal Retrieval ($L \to V$):} This paradigm enables "blind" localization, where a system recognizes a place from a textual description alone. This capability is critical for semantic robotics (interpreting human instructions), forensic geolocation (witness descriptions), and emergency rescue calls. For instance, a distressed caller might describe high-stress conditions, e.g., \textit{"someone fainted and I am near a tall stone building with a clock tower"}, requiring the system to geolocate the scene solely through linguistic cues.
This task requires learning a shared latent space that can align abstract verbal cues with the specific, fine-grained visual features necessary for precise localization.
\end{enumerate}
While these paradigms offer a path toward more resilient localization, mainstream VPR benchmarks are image-only, lacking the textual descriptions required to train and evaluate vision-language models. To address this, we introduce \textit{LaVPR}, a large-scale benchmark comprising over 650,000 descriptions. LaVPR systematically extends established VPR datasets with linguistic data generated via a Visual Large Language Model (VLLM). To ensure spatial fidelity and reduce hallucinations, we employ a segmentation-grounded pipeline and a human-in-the-loop protocol (Section~\ref{sec:dataset}).

Using LaVPR, we provide a comprehensive evaluation of vision-language strategies. For Multi-Modal Fusion (Section~\ref{sec:multimodal}), our results reveal consistent performance gains across a diverse range of visual backbones. Moreover, we find that incorporating linguistic context allows compact visual backbones (e.g., ViT-S) to rival the accuracy of much larger, compute-intensive models. This offers a path toward high-throughput, efficient deployment. For Cross-Modal Retrieval (Section~\ref{sec:crossmodal}), we observe that standard zero-shot transfer and contrastive fine-tuning fail to achieve precise localization. We instead propose a parameter-efficient alignment strategy using Low-Rank Adaptation (LoRA)~\cite{hu2022lora} in conjunction with the Multi-Similarity (MS) loss~\cite{wang2019multi}, establishing a robust baseline for cross-modal VPR. 

In summary, our contributions are as follows:
\begin{itemize}
\item We introduce LaVPR, a standardized, open-source benchmark for vision-language place recognition that extends established VPR datasets with over 650,000 aligned, grounded textual descriptions.
\item We demonstrate that language descriptions provide consistent accuracy gains across diverse architectures, showing that linguistic context allows compact backbones to rival the performance of significantly larger vision-only models.
\item We present an empirical study on cross-modal alignment for ``blind'' localization, demonstrating that pairing LoRA with the MS loss is critical for learning a discriminative shared embedding space, substantially outperforming standard contrastive baselines.
\end{itemize}

\section{Related Work}\label{sec:related_work}
\subsection{Visual Place Recognition} 
VPR systems are defined by three primary design choices: the architectural backbone, the aggregation strategy, and the training objective. Backbones have transitioned from standard CNNs~\cite{arandjelovic2016netvlad,radenovic2018fine} to Transformer-based models that leverage self-attention for superior global context~\cite{lu2024cricavpr}. To compress these features into a global descriptor, methods utilize aggregation strategies such as NetVLAD’s residual clustering~\cite{arandjelovic2016netvlad} or Generalized Mean (GeM) pooling~\cite{radenovic2018fine}. Recent gains involve enhancing these descriptors through attention mechanisms~\cite{huang2024dino}, multi-scale aggregation~\cite{hausler2021patch,ali2023mixvpr}, and adapter-based fine-tuning of foundation models~\cite{lu2024towards}. This shift reflects a trend where pre-trained, large-scale encoders provide robust features that often outperform specialized, task-specific architectures. Training typically utilizes contrastive objectives, such as triplet~\cite{hoffer2015deep} or multi-similarity loss~\cite{wang2019multi}, to enforce invariance to environmental changes while maximizing spatial separation. To allow training on dense city-scale datasets, recent methods adopt classification formulations~\cite{berton2022rethinking,berton2023eigenplaces} utilizing large-margin cosine losses~\cite{wang2018cosface}.

The field relies on retrieval metrics (Recall$@K$) across benchmarks characterized by viewpoint shifts and extreme temporal variations~\cite{amstertime,pitts,torii201524}. Large-scale training sets like GSV-Cities~\cite{Alibey_2022_gsvcities} and SF-XL~\cite{berton2022rethinking} have become standard for these tasks. LaVPR builds on these foundations by augmenting these widely-used datasets with over 650,000 aligned textual descriptions. Unlike previous works limited to uni-modal evaluation, this extension enables a controlled benchmark of multi-modal and cross-modal strategies using identical visual backbones, isolating the impact of linguistic context.
\subsection{Language-Vision Place Recognition} 
The exploitation of complementary sensor inputs, and language in particular, remains underexplored for VPR research~\cite{li2025place}.
\paragraph{Multi-Modal Place Recognition.} Prior fusion methods have predominantly integrated 3D geometry (depth or LiDAR) with imagery~\cite{hu2020dasgil,piasco2021improving,li2025unified,komorowski2021minkloc++,lai2022adafusion,zhou2023lcpr}, consistently improving robustness over vision-only approaches. Within the text domain, methods like TextPlace~\cite{hong2019textplace} have leveraged scene text through Optical Character Recognition (OCR) as an invariant feature, capitalizing on the distinctiveness of street signs and shop names for location identification.

Most recently, MSSPlace~\cite{melekhin2025mssplace} explored fusing diverse data streams, including multi-camera imagery, LiDAR, and text, into a unified global descriptor. To facilitate this, the authors extended the Oxford RobotCar~\cite{maddern20171} and NCLT~\cite{ncarlevaris-2015a} datasets using MiniGPT-4~\cite{zhu2024minigpt}. While this work provided valuable evidence for the benefits of multi-modal fusion, its evaluation protocol lacked a controlled ablation, comparing multi-modal, multi-view inputs against single-view baselines trained on disparate datasets. Furthermore, the absence of a released dataset (to date) or a detailed curation protocol limits the reproducibility of their findings. Our work addresses these limitations by introducing a rigorous, publicly available framework for standardized language-vision place encoding. By maintaining a strictly controlled experimental setup, we isolate the specific contribution of language, revealing new insights into how semantic descriptions complement visual features to achieve robust VPR.
\paragraph{Cross-Modal Place Recognition.} Research in this area has primarily focused on cross-modal text-to-point-cloud localization~\cite{xia2024text2loc,xu2025cmmloc,liu2025text}. Introduced by Kolmet et al.~\cite{kolmet2022text2pos}, this task aims to regress a 6-DoF pose within a 3D environment using only natural language descriptions. Standard pipelines typically employ a coarse-to-fine approach: mapping textual queries and 3D sub-maps into a shared embedding space for retrieval, followed by fine-grained matching of semantic instances. Evaluation relies on datasets like KITTI360Pose~\cite{kolmet2022text2pos}, an extension of KITTI360~\cite{xie2016semantic} containing template-based descriptions of spatial relations. In the text-to-image domain, Shang et al. recently proposed a multi-view registration approach to align grouped textual descriptions with visual scenes~\cite{shang2025bridging}. However, this method diverges from the standard VPR paradigm by requiring multi-view textual queries and focusing on set-to-set alignment rather than scalable, single-query global descriptor retrieval on established benchmarks. In LaVPR, we directly extend VPR training and evaluation benchmarks with textual descriptions, curate them and then analyze the contribution in a controlled setting.

\section{The LaVPR Benchmark}\label{sec:dataset}
We introduce LaVPR, a large-scale benchmark designed to bridge visual localization and natural language. By augmenting established VPR datasets with 651,865 aligned descriptions (Table~\ref{tab:datasets}), LaVPR enables the systematic evaluation of multi-modal fusion and cross-modal retrieval while preserving original geographic splits for fair comparison.
\subsection{LaVPR Generation}
We employ Gemini 2.5 Flash~\cite{google2025gemini} to generate rich, high-density narratives that prioritize permanent architectural features and spatial relationships over transient elements like pedestrians.
\begin{table*}[ht!]
  \centering
  \scriptsize
  \caption{Overview of the VPR datasets integrated into {LaVPR}. The combined dataset consists of {651,865} image-description pairs. Dataset characteristics are curated based on \cite{lu2024cricavpr}.}
  \label{tab:datasets}
  \renewcommand{\arraystretch}{1.2}
  \begin{tabular}{ll c rr}
    \toprule
    \multirow{2}{*}{\textbf{Dataset}} & \multirow{2}{*}{\textbf{Type}} & \multirow{2}{*}{\textbf{Usage}} & \multicolumn{2}{c}{\textbf{Image Count}} \\
    \cmidrule(lr){4-5} 
    & & & \textbf{Database} & \textbf{Queries} \\ 
    \midrule
    GSV-Cities~\cite{Alibey_2022_gsvcities} & Urban & Train & 529,506 & -- \\ 
    \midrule
    Pitts30K-Val~\cite{pitts} & Urban, Panorama & Validation & 10,000 & 7,608 \\ 
    Pitts30K-Test~\cite{pitts} & Urban, Panorama & Test & 10,000 & 6,816 \\ 
    AmsterTime~\cite{amstertime} & Very long-term & Test & 1,231 & 1,231 \\ 
    MSLS-val~\cite{msls} & Urban, Suburban & Test & 18,871 & 740 \\ 
    MSLS-challenge~\cite{msls} & Long-term & Test & 38,770 & 27,092 \\ 
    \midrule
    \rowcolor{gray!10} \textbf{Total (LaVPR)} & & & \textbf{608,378} & \textbf{43,487} \\
    \bottomrule
  \end{tabular}
\end{table*}

\begin{figure}[h!]
    \centering
\includegraphics[scale=0.35]{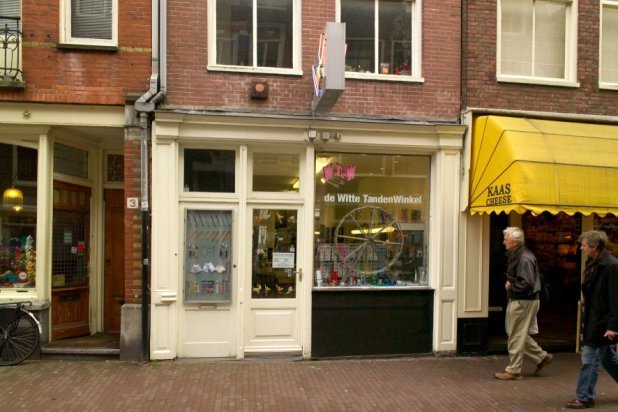}
    \caption{Example image from the Amstertime dataset. {Aligned description in LaVPR:} 'Leftmost red brick building with an upper window and a black wrought iron balcony railing, cream-colored framed storefront with a display window, recessed wooden entrance door, narrow brick wall section with house number '3', grey vertical downspout, middle cream-colored framed shop with a glass door, large storefront window displaying "de Witte TandenWinkel", black lower facade panel, red brick upper floor with a rectangular window, blue square sign, grey electrical box, rightmost red brick building with an upper window, yellow scalloped awning labeled "KAAS CHEESE".'}
\label{fig:amstertime_example}
\end{figure}

Unlike sparse keyword-based datasets, LaVPR provides dense semantic descriptions (Figure~\ref{fig:amstertime_example}) that capture fine-grained textures and scene text. This descriptive depth is summarized in Table~\ref{tab:lavpr_statistics}.
To isolate the impact of language when vision is compromised, we further curated specialized subsets targeting visual degradation and discriminative scene text (detailed in Table~\ref{tab:challenging_datasets}). Section 1 in our Supplementary (Suppl.) Material provides the generation prompt and the specialized dataset generation protocols, including example images.
\begin{table}[h!]
  \centering
  \scriptsize
\caption{Specialized subsets in LaVPR. These datasets target high linguistic discriminativeness and extreme visual domain gaps to isolate the impact of the language channel.}
  \label{tab:challenging_datasets}
  \renewcommand{\arraystretch}{1.2}
  \setlength{\tabcolsep}{3pt}
  \begin{tabular}{lll r}
    \toprule
    \textbf{Dataset} & \textbf{Key Query} & \textbf{Reference} & \textbf{\# Queries} \\
         & \textbf{Properties} & \textbf{Database} &  \\
    \midrule
    Amstertime-La & Scene text & Amstertime & 31 \\
    MSLS-Blur & Scene text, blur & MSLS-val & 86 \\
    MSLS-Weather & Scene text, adverse weather & MSLS-val & 86 \\
    \bottomrule
  \end{tabular}
\end{table}
\begin{table}[tbh]
  \centering
  \scriptsize
  \caption{Key linguistic and structural statistics of the {LaVPR} benchmark. These metrics highlight the descriptive density and lexical diversity across the integrated datasets.}
  \label{tab:lavpr_statistics}
  \renewcommand{\arraystretch}{1.2}
  \setlength{\tabcolsep}{3pt}
  \begin{tabular}{l cccc}
    \toprule
    \textbf{Metric} & \textbf{Training} & \textbf{Validation} & \textbf{Test} & \textbf{Total} \\
    \midrule
    Total Image-Text Pairs & 529,506 & 17,608 & 104,751 & \textbf{651,865} \\
    Avg. Word Count per Caption & 52.78 & 50.35 & 46.43 & 51.69 \\
    Avg. Entity Count (Nouns) & 23.66 & 22.49 & 21.35 & 23.25 \\
    Total Unique Vocabulary & 176,893 & 16,455 & 55,110 & 209,200 \\
    Signage/Text-heavy Samples (\%) & 27.77\% & 55.66\% & 30.06\% & 28.89\% \\
    \bottomrule
  \end{tabular}
\end{table}
\subsection{LaVPR Curation}

While modern VLMs generate rich descriptors, they remain susceptible to hallucinations~\cite{huang2025survey}. To rigorously assess and ensure data reliability, we developed a comprehensive validation pipeline that integrates automated \textit{entity extraction} (Phi-3.5-mini-instruct\footnote{\scriptsize \url{https://huggingface.co/microsoft/Phi-3.5-mini-instruct}, ~\cite{abdin2024phi-3}}), \textit{spatial grounding} (SAM3\footnote{\scriptsize \url{https://huggingface.co/facebook/sam3},~\cite{carion2025sam}}), and \textit{binary VLM verification} (Qwen2-VL-7B-Instruct\footnote{\scriptsize \url{https://huggingface.co/Qwen/Qwen2-VL-7B-Instruct},~\cite{wang2024qwen2}}). To maximize safety, the automated stage is tuned for exceptionally high recall (91\%) over precision (22\%), as measured on a manually curated validation set. This stringent tuning ensures that virtually all potential discrepancies are flagged for review, albeit with a high rate of false positives. We resolve these ambiguities through a Human-in-the-Loop (HL) phase, distinguishing genuine hallucinations from false alarms caused by out-of-distribution elements or visual noise.

We deployed this auditing protocol across our evaluation splits to verify benchmark fidelity. Crucially, while the hyper-sensitive automated filter flagged 26\% of samples for secondary verification, subsequent HL analysis of the largest unresolved groups revealed that the raw data is inherently of high quality: genuine hallucinations were  rare, affecting only $\sim 1\%$ of the inspected images. Typically, these minor cases involved a single erroneous object within otherwise precise descriptions; these rare confirmed hallucinations were subsequently purged from the evaluation sets. Furthermore, manual validation of in-scene text confirmed a 93.2\% accuracy rate, with minor errors isolated to severe occlusions. 

We emphasize that all main-paper results are obtained using the raw, uncurated training data. The curation pipeline described above was used purely as a diagnostic audit to guarantee evaluation integrity, rather than as a necessary cleaning step for model optimization. Crucially, achieving strong performance on raw, VLM-generated data demonstrates both the high baseline quality of LaVPR and our framework's inherent robustness to minor visual-language noise. For completeness, a comprehensive algorithmic walkthrough of the curation pipeline, step-by-step statistical breakdowns, and an ablation study of our cleaning strategy are provided in Section 2 of the Suppl. Material.
\section{Multi-Modal Place Recognition ($V+L$)}\label{sec:multimodal}
We investigate the synergy between natural language and vision to address the inherent fragility of image-only descriptors. By integrating linguistic context, we provide a stable, high-level semantic reference that persists across temporal and environmental shifts.
\subsection{Late-Fusion Methods}
We evaluate four late-fusion architectures utilizing frozen, pretrained visual ($E_v$) and textual ($E_t$) encoders. This design isolates cross-modal dynamics and demonstrates the potential of language as a "robustness anchor" without requiring extensive end-to-end retraining.

Given an image $I$ and description $t$, we obtain embeddings $z_v = E_v(I) \in \mathbb{R}^{d_v}$ and $z_t = E_t(t) \in \mathbb{R}^{d_t}$. We explore the following fusion operations:
\begin{itemize}
    \item Concatenation (CAT): A baseline where $z_e = [z_v; z_t] \in \mathbb{R}^{d_v+d_t}$.
    \item Projection and Addition (PA): Linear layers project both modalities to a shared dimension $d_e$, where $z_e = \text{Linear}(z_v) + \text{Linear}(z_t)$.
    \item Multi-Layer Perceptron (MLP): A shallow MLP processes the concatenated vector, expanding the dimension to $(d_v+d_t)$ before mapping to a fixed latent space $d_e$.
    \item Adaptive Score Fusion (ADS): Instead of feature fusion, we propose to adaptively weigh modality-specific similarity scores (depicted in Fig. 10 in the Suppl. Material). An MLP with Softmax predicts weights $W=[w_v, w_t]$ from concatenated visual and language query features. For a query-reference pair $(i, j)$, weights are averaged: $w_{m_{ij}} = (w_{m_i} + w_{m_j})/2$ for $m \in \{v, t\}$. The joint similarity is $S_{ij} = w_{v_{ij}} \cdot S_{v_{ij}} + w_{t_{ij}} \cdot S_{t_{ij}}$.
\end{itemize}
All fusion modules, except for CAT,  are optimized using the Multi-Similarity (MS) loss~\cite{wang2019multi}:
\begin{equation}
	\label{eq:MS_Loss}
    \begin{split}
	\mathcal{L}_{MS} = \frac{1}{|B|}\sum_{q  \in \mathcal{B}}  \bigg\{\frac{1}{\alpha}  { \log \big[1 + \sum_{p  \in \mathcal{P}_q } e^{-\alpha (S_{qp} - \lambda)}}\big]  \\
	+ \frac{1}{\beta }  { \log \big[1+ \sum_{n \in \mathcal{N}_q}
		 e^{\beta (S_{qn} - \lambda)} \big]} \bigg\},
         \end{split}
\end{equation}
where for each query $q$ in a batch $B$, $\mathcal{P}_q$ and $\mathcal{N}_q$ correspond to the sets of positive and negative samples for $q$, respectively, and $S_{qp}$ and $S_{qn}$ are the cosine similarities of a positive pair $\{q,p\}$ and a negative pair $\{q,n\}$, correspondingly. $\alpha$ and $\beta$ and $\lambda$ are hyper-parameters. In ADS, we use the multi-modal similarity scores as input to the MS loss.

To distill salient linguistic features, we further introduce a Learned Language Pooling (LLP) module. This module employs a self-attention mechanism over the hidden states of $E_t$ to generate a context-aware textual representation. We evaluate this pooling mechanism in combination with the four late fusion strategies. Additional implementation details are provided in Section 5 in the Suppl. Material.
\subsection{Results \& Discussion}
We evaluate multi-modal fusion across diverse VPR models, including NetVLAD~\cite{arandjelovic2016netvlad}, CosPlace~\cite{berton2022rethinking}, EigenPlaces~\cite{berton2023eigenplaces}, MixVPR~\cite{ali2023mixvpr}, SALAD~\cite{izquierdo2024optimal}, and CricaVPR~\cite{lu2024cricavpr}, using BAAI/bge-large-en-v1.5 (BGE-L) model \cite{bge_embedding} as the primary text encoder.
\paragraph{Evaluation of Fusion Mechanisms:} 
As shown in Table~\ref{tab:compare_fusion_mechanism}, ADS and CAT emerge as the most effective multi-modal integration strategies when paired with MixVPR. While CAT achieves competitive performance across datasets, ADS yields peak robustness when paired with LLP. For example, ADS with LLP achieves the highest absolute performance on Amstertime, improving R@1 to 38.1\%. Conversely, CAT performance degrades  when combined with LLP, indicating that fixed structural concatenation cannot accommodate independently projected modalities without dynamic rescaling, which ADS naturally handles via adaptive scoring. We further show that CAT and ADS provide consistent gains across different textual encoders in our Suppl. Material (Section 4, Table 4).

\begin{table*}[h!]
\centering
\scriptsize
\caption{Comparison of late-fusion mechanisms using MixVPR (embedding dimension of 512) and BGE-L as the visual and textual backbones, respectively. Shading indicates the \colorbox{best}{best} and \colorbox{runnerup}{runner-up} multi-modal configurations per dataset.} 
\label{tab:compare_fusion_mechanism}
\renewcommand{\arraystretch}{1.2}
\begin{tabular}{l ccc ccc ccc}
\toprule
\textbf{Fusion} & \multicolumn{3}{c}{\textbf{Amstertime}} & \multicolumn{3}{c}{\textbf{MSLS-val}} & \multicolumn{3}{c}{\textbf{Pitts30}} \\
\cmidrule(lr){2-4} \cmidrule(lr){5-7} \cmidrule(lr){8-10}
\textbf{Mechanism} & R@1 & R@5 & R@10 & R@1 & R@5 & R@10 & R@1 & R@5 & R@10 \\
\midrule
Visual-only & 35.7 & 53.1 & 60.4 & 83.2 & 91.9 & 93.4 & 90.6 & 95.6 & 96.3 \\
\midrule
\rowcolor{groupgray} \multicolumn{10}{l}{\textit{Without LLP}} \\
ADS & 36.1 & 53.4 & 60.5 & \cellcolor{best} \textbf{83.4} & \cellcolor{best} \textbf{91.9} & \cellcolor{runnerup} 93.4 & \cellcolor{runnerup} 90.7 & \cellcolor{best} \textbf{95.6} & 96.3 \\
CAT & \cellcolor{runnerup} 37.0 & \cellcolor{runnerup} 58.2 & \cellcolor{best} \textbf{65.6} & 81.9 & 91.5 & 93.0 & 89.9 & 95.3 & \cellcolor{best} \textbf{96.4} \\
MLP & 31.7 & 52.6 & 59.8 & 83.2 & 91.5 & 93.0 & 89.8 & 95.1 & 96.2 \\
PA  & 31.0 & 50.8 & 57.3 & 83.0 & 90.1 & 91.9 & 89.7 & 95.1 & 96.3 \\
\midrule
\rowcolor{groupgray} \multicolumn{10}{l}{\textit{With LLP}} \\
ADS & \cellcolor{best} \textbf{38.1} & \cellcolor{best} \textbf{58.4} & \cellcolor{runnerup} 64.7 & \cellcolor{runnerup} 83.2 & 91.6 & \cellcolor{best} \textbf{93.5} & \cellcolor{best} \textbf{90.8} & \cellcolor{runnerup} 95.4 & \cellcolor{best} \textbf{96.4} \\
CAT & 30.2 & 51.0 & 59.7 & 77.4 & 89.1 & 91.6 & 85.7 & 93.9 & 95.8 \\
MLP & 32.6 & 50.8 & 58.3 & \cellcolor{best} \textbf{83.4} & 90.5 & 92.8 & 89.6 & 95.1 & 96.2 \\
PA  & 30.2 & 47.5 & 54.9 & 82.3 & 90.1 & 92.3 & 89.6 & 94.8 & 96.1 \\
\bottomrule
\end{tabular}
\end{table*}

\paragraph{Influence of Pre-training Paradigms and Backbone Architecture:} 
Beyond the fusion mechanism itself, the magnitude of linguistic gains is heavily governed by the architecture of the visual backbone and its pre-training objective (Table~\ref{tab:relative_gains_pretraining}). Supervised convolutional backbones (VGG and ResNet) exhibit the most dramatic relative improvements, with NetVLAD achieving up to a 200\% R@1 gain, indicating that classification-based pre-training leaves substantial "semantic gaps" regarding environmental context that text effectively bridges. 

Conversely, self-supervised (SSL) Vision Transformer backbones like DINOv2 capture a much richer visual-semantic prior, naturally narrowing the margin for linguistic improvement ($<7\%$). Yet, even within this saturated regime, our proposed ADS+LLP fusion framework yields consistent, non-trivial advantages where simple CAT fusion plateaus. 
\begin{table*}[t]
\centering
\scriptsize
\caption{Relative R@1 Gain (\%) of Language-Augmented (La-) methods compared to visual-only baselines. \textcolor{posgain}{Positive gains} indicate linguistic complementarity; \textcolor{negloss}{negative values} indicate slight degradation. Embedding dimension for each backbone is denoted in brackets ($d$). VGG16 and ResNet50 backbones are pretrained on ImageNet-1k in a supervised manner. DinoV2-L is pretrained via invariance-based self supervised learning (SSL). 
VPR Loss Abbreviations: {TRP}: Triplet Loss; {LMC}: Large Margin Cosine Loss; {MS}: Multi-Similarity Loss.}
\label{tab:relative_gains_pretraining}
\renewcommand{\arraystretch}{1.5}
\begin{tabular}{lll c ccccc}
\toprule
\textbf{Method} & \textbf{Fusion} & \textbf{Backbone ($d$)} & \textbf{Loss} & \multicolumn{5}{c}{\textbf{Relative R@1 Gain (\%) per Dataset}} \\
\cmidrule(lr){5-9}
& & & & \textbf{MSLS-B} & \textbf{MSLS-W} & \textbf{Ams.} & \textbf{MSLS-C} & \textbf{Pitts30} \\
\midrule
La-NetVLAD    & CAT & VGG16 (32k)    & TRP & \textcolor{posgain}{+162.3} & \textcolor{posgain}{+136.3} & \textcolor{posgain}{+199.5} & \textcolor{posgain}{+18.3} & \textcolor{negloss}{-1.0} \\
\midrule
La-CosPlace   & CAT & ResNet50 (512) & LMC & \textcolor{posgain}{+44.9}  & \textcolor{posgain}{+29.4}  & \textcolor{posgain}{+13.2}  & \textcolor{posgain}{+5.7}  & \textcolor{posgain}{+0.3} \\
La-EigenPlace & CAT & ResNet50 (2k)  & LMC & \textcolor{posgain}{+43.2}  & \textcolor{posgain}{+15.3}  & \textcolor{posgain}{+15.5}  & \textcolor{posgain}{+3.4}  & \textcolor{posgain}{+0.2} \\
\midrule
La-MixVPR     & CAT & ResNet50 (4k)  & MS  & \textcolor{posgain}{+21.3}  & \textcolor{posgain}{+14.9}  & \textcolor{posgain}{+6.0}   & \textcolor{posgain}{+0.9}  & \textcolor{negloss}{-0.9} \\
La-MixVPR     & ADS+LLP & ResNet50 (4k)  & MS  & \textcolor{posgain}{+14.8}  & \textcolor{posgain}{+10.4}  & \textcolor{posgain}{+11.9}  & \textcolor{posgain}{+2.2}  & \textcolor{posgain}{+0.3} \\
\midrule
La-SALAD      & CAT & DinoV2-L (8k)  & MS  & \textcolor{posgain}{+1.2}   & \textcolor{posgain}{+3.9}   & \textcolor{posgain}{+5.5}   & \textcolor{negloss}{-0.9}  & \textcolor{negloss}{-0.8} \\
La-SALAD      & ADS+LLP & DinoV2-L (8k)  & MS  & \textcolor{posgain}{+1.2}   & \textcolor{posgain}{+3.9}   & \textcolor{posgain}{+5.5}   & 0.0   & \textcolor{posgain}{+0.1} \\
\midrule
La-CricaVPR   & CAT & DinoV2-L (10k) & MS  & 0.0     & \textcolor{posgain}{+5.2}   & \textcolor{negloss}{-9.9}   & \textcolor{negloss}{-3.7}  & \textcolor{negloss}{-1.5} \\
La-CricaVPR   & ADS+LLP & DinoV2-L (10k) & MS  & \textcolor{posgain}{+1.2}   & \textcolor{posgain}{+6.6}   & 0.0    & \textcolor{posgain}{+0.6}  & \textcolor{posgain}{+0.1} \\
\bottomrule
\end{tabular}
\end{table*}

\begin{figure}[h!]
    \centering
    \includegraphics[scale=0.3]{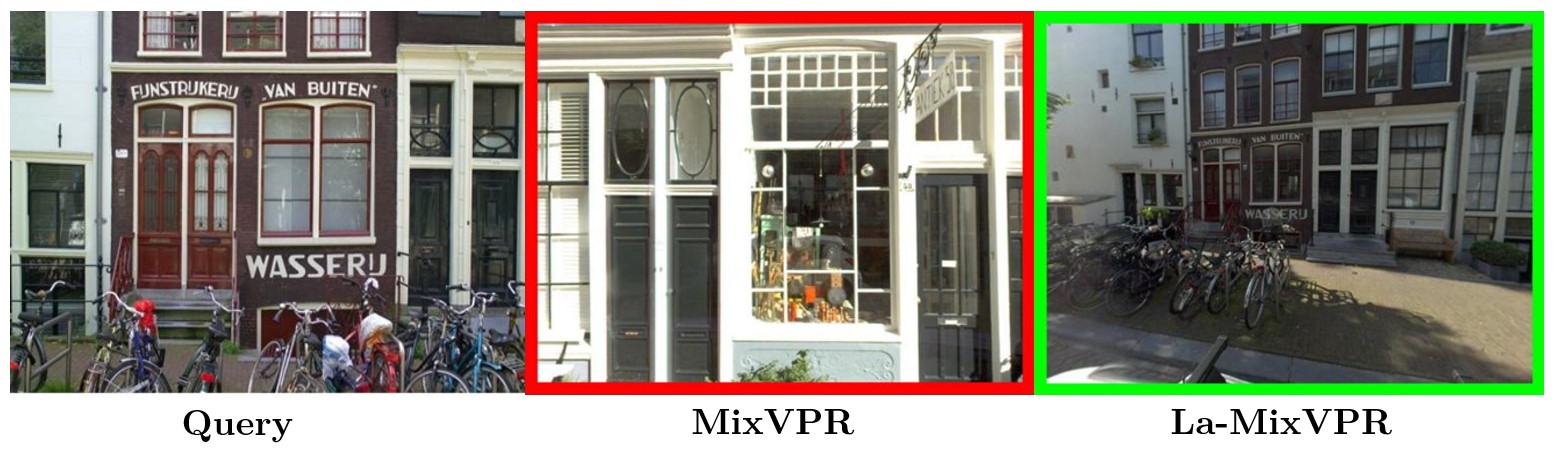}
    \caption{Qualitative results from the Amstertime dataset. From left to right: Query, MixVPR Top-1, and La-MixVPR Top-1 (using CAT Fusion). Correct matches are bordered in green; incorrect in red. Additional qualitative results are provided in the Suppl. Material.}
    \label{fig:comparison}
\end{figure}

\paragraph{Compensatory Dynamics in Adverse Environments:} 
The utility of language as a compensatory signal becomes most pronounced under adverse conditions where visual reliability is compromised (Table~\ref{tab:lavpr_multimodal_ads}). Language-augmented (La-) methods consistently outperform their visual-only baselines for evaluation subsets characterized by extreme blur and severe weather. In these degraded scenarios, the textual description functions as a robust semantic anchor that remains completely invariant to pixel-level visual domain shifts. A qualitative example illustrating how La-MixVPR successfully resolves an environmental degradation that misleads baseline MixVPR is shown in Fig.~\ref{fig:comparison}.
\begin{table*}[h!]
\centering
\scriptsize     
\caption{Comparison of vision-only VPR methods vs. our Language-Augmented (La-) variants using {ADS+LLP Fusion} and BGE-L. Bold indicates the best performance within each backbone pair.}
\label{tab:lavpr_multimodal_ads}
\renewcommand{\arraystretch}{1.5}
\setlength{\tabcolsep}{1.5pt}
\begin{tabular}{lr ccc ccc ccc ccc}
\toprule
\textbf{Method} & \textbf{Dim} & \multicolumn{3}{c}{\textbf{MSLS-B}} & \multicolumn{3}{c}{\textbf{MSLS-W}} & \multicolumn{3}{c}{\textbf{Amst.-La}} & \multicolumn{3}{c}{\textbf{MSLS Ch.}}  \\
\cmidrule(lr){3-5} \cmidrule(lr){6-8} \cmidrule(lr){9-11} \cmidrule(lr){12-14}
& ($d$) & R@1 & R@5 & R@10 & R@1 & R@5 & R@10 & R@1 & R@5 & R@10 & R@1 & R@5 & R@10 \\
\midrule
MixVPR & 4096 & 70.9 & 82.6 & 87.2 & 77.9 & 83.7 & 88.4 & 54.8 & 80.6 & 83.9 & 63.4 & 76.0 & 79.2  \\
\rowcolor{highlight} La-MixVPR & +1024 & \textbf{81.4} & \textbf{93.0} & \textbf{97.7} & \textbf{86.0} & \textbf{91.9} & \textbf{96.5} & \textbf{61.3} & \textbf{90.3} & \textbf{90.3} & \textbf{64.8} & \textbf{76.9} & \textbf{80.3}  \\
\midrule
SALAD & 8448 & 91.9 & 98.8 & 98.8 & 90.7 & 96.5 & 97.7 & 58.1 & 90.3 & 93.5 & 74.6 & 88.4 & 91.0 \\
\rowcolor{highlight} La-SALAD & +1024 & \textbf{93.0} & 98.8 & 98.8 & \textbf{94.2} & \textbf{98.8} & \textbf{100} & \textbf{61.3} & 90.3 & 93.5 & \textbf{74.6} & \textbf{88.8} & \textbf{91.2} \\
\midrule
CricaVPR & 10752 & 91.9 & \textbf{98.8} & 98.8 & 88.4 & 94.2 & 97.7 & 64.5 & 90.3 & 93.5 & 69.8 & \textbf{82.2} & 85.4 \\
\rowcolor{highlight} La-CricaVPR & +1024 & \textbf{93.0} & \textbf{98.8} & \textbf{98.8} & \textbf{94.2} & \textbf{98.8} & \textbf{98.8} & 64.5 & \textbf{96.8} & \textbf{96.8} & \textbf{70.2} & 82.1 & \textbf{86.1}  \\
\bottomrule
\end{tabular}
\end{table*}. 

A similar trend is observed also when utilizing the CAT fusion (Suppl. Material, Section 4, Table 3). However, while CAT is competitive for compact descriptors, the ADS-LLP framework provides greater scalability and robustness for high-dimensional backbones like CricaVPR.

\paragraph{Computational Efficiency and Scaling:} Language integration further offers a path toward efficient deployment. As shown in Table~\ref{tab:complexity}, by pairing a compact visual backbone (CricaVPR with ViT-Small) with a lightweight text encoder (BGE-Small), we achieve 93.0 R@1 on MSLS-Blur, outperforming the massive visual-only CricaVPR-Base. This hybrid approach reduces computational complexity by 62\% (6.82 GFLOPs vs. 17.7 GFLOPs), demonstrating that horizontal scaling via multi-modality is a more efficient route to robustness than the vertical scaling of transformers 
architectures.
\begin{table}[h!]
\centering
\scriptsize
\caption{Efficiency-Performance Trade-off. We compare current state-of-the-art (SOTA) VPR, namely, CricaVPR, and La-CricaVPR (ours) with smaller backbones and CAT fusion. CricaVPR is a state-of-the-art (SOTA) VPR model with a Vit-L backbone. CricaVPR-Small corresponds to CricaVPR with a ViT-S backbone. La-CricaVPR-Small pairs CricaVPR-Small with BGE-Small (textual encoder). It achieves higher R@1 than the base SOTA model with $\sim 62\%$ fewer total GFLOPs. Abbreviations for complexity (GFLOPs): Inf.: GFLOPs count for embedding the query (Inference); Retr.: GFLOPs count for retrieval against the reference database (Retrieval); Total is the sum of Inf. and Retr.}
\label{tab:complexity}
\renewcommand{\arraystretch}{1.5}
\begin{tabular}{lc ccc c cc cc cc}
\toprule
\textbf{Configuration} & \textbf{Params} & \multicolumn{3}{c}{\textbf{GFLOPs}} & \textbf{Dim} & \multicolumn{2}{c}{\textbf{MSLS-B}} & \multicolumn{2}{c}{\textbf{MSLS-W}} & \multicolumn{2}{c}{\textbf{Amst.-La}} \\
\cmidrule(lr){3-5} \cmidrule(lr){7-8} \cmidrule(lr){9-10} \cmidrule(lr){11-12}
& (M) & Inf. & Retr. & {Total} & ($d$) & R@1 & R@5 & R@1 & R@5 & R@1 & R@5 \\
\midrule
CricaVPR & 106.8 & 17.5 & 0.20 & 17.7 & 10k & 91.9 & \textbf{98.8} & 88.4 & 94.2 & \textbf{64.5} & 90.3 \\
CricaVPR-Small      & 27.2 & 4.6 & 0.10 & 4.8 & 5k & 89.5 & 95.3 & 84.9 & 91.9 & 54.8 & 87.1 \\
\midrule
\textbf{La-Crica-Small} & 60.6 & 6.7 & 0.12 & 6.82 & 6k & \textbf{93.0} & 97.7 & \textbf{90.7} & \textbf{96.5} & 58.1 & \textbf{93.5} \\
\bottomrule

\end{tabular}
\end{table}

\paragraph{Complementary Synergy versus Sequential Retrieval:} Finally, we compare our joint approach against sequential re-ranking (Top-100 retrieval followed by cross-modal re-ranking). As shown in Table~\ref{tab:rerank_comparison}, sequential ranking fails catastrophically (R@1 drops to 1.2\%) due to the retrieval bottleneck: if the primary modality misses the ground truth in the initial top-$K$, the second cannot recover it. Furthermore, candidates that rank highly in one modality do not necessarily possess discriminative features in the other.  Joint fusion avoids this by allowing both modalities to mutually disambiguate the search space from the start.
\begin{table}[h!]
  \centering
\caption{Comparison of Joint Fusion (Ours) vs. Sequential Re-ranking on Amstertime. Sequential methods re-rank the top-100 candidates from the initial search modality.} 
\label{tab:rerank_comparison}
  \scriptsize  \renewcommand{\arraystretch}{1.2}
  \begin{tabular}{ll cccc}
  \toprule
  \textbf{Initial Search} & \textbf{Re-ranked by} & \multicolumn{4}{c}{\textbf{Amstertime (Recall@K)}} \\
  \cmidrule(lr){3-6}
  & & R@1 & R@5 & R@10 & R@20 \\
  \midrule
  \rowcolor{gray!10} \multicolumn{6}{l}{\textit{Single Modality Baselines}} \\
  MixVPR (Visual) & -- & 35.7 & 53.1 & 60.4 & 65.9 \\
  BGE (Textual)   & -- & 9.5  & 18.0 & 23.5 & 32.4 \\
  \midrule
  \rowcolor{gray!10} \multicolumn{6}{l}{\textit{Sequential Re-ranking (Top-100)}} \\
  MixVPR          & BGE (Text)    & 1.2  & 3.4  & 5.2  & 6.6  \\
  BGE             & MixVPR (Vis)  & 4.4  & 5.6  & 5.9  & 6.3  \\
  \midrule
  \rowcolor{gray!10} \multicolumn{6}{l}{\textit{Joint Multi-modal Fusion}} \\
  \textbf{LaVPR (Ours)} & \textbf{--} & \textbf{37.0} & \textbf{58.2} & \textbf{65.6} & \textbf{71.6} \\
  \bottomrule
  \end{tabular}
\end{table}
\section{Cross-Modal Place Recognition ($L \rightarrow V$)}\label{sec:crossmodal}
While multi-modal fusion leverages language as a robust auxiliary signal, a more fundamental challenge lies in cross-modal retrieval: identifying a geographic location within a visual database using only a natural language query. This task requires alignment between abstract linguistic descriptions and the concrete, structural visual features that define a specific place.
\subsection{Aligning Language and Vision for 'Blind' Localization}
Given an image $I$ and a textual description $t$, let $E_v$ and $E_t$ denote the visual and textual encoders mapping $I$ and $t$ to embeddings $\mathbf{z}_v \in \mathbb{R}^{d_v}$ and $\mathbf{z}_t \in \mathbb{R}^{d_t}$, respectively. The objective is to retrieve the ground-truth image $I^*$ from a database $\mathcal{D}$ that corresponds to a query description $t_q$.
Note that the text description generated from a query image are is solely to retrieve geographically matching, yet visually distinct, reference images from the database, while the query image itself never appears in the search database.
Standard vision-language models are pre-trained to align object-level semantics rather than the fine-grained spatial and structural cues required to discriminate between visually similar locations. To bridge this gap, we do not learn a new latent space from scratch; instead, we re-orient the existing shared embedding space toward place-discriminative features using Low-Rank Adaptation (LoRA)~\cite{hu2022lora}. For a pre-trained weight matrix $W_0 \in \mathbb{R}^{d \times k}$, the weight update is represented as a low-rank decomposition:
\begin{equation}
    W = W_0 + \Delta W = W_0 + BA
\end{equation}
where $B \in \mathbb{R}^{d \times r}$ and $A \in \mathbb{R}^{r \times k}$, with rank $r \ll \min(d, k)$. We apply these updates to all linear projections within the Transformer layers of our vision-language backbones. The model is optimized using the Multi-Similarity (MS) loss (Eq.~\ref{eq:MS_Loss}), which encourages a discriminative cross-modal manifold by mining hard positive and negative pairs across modalities. We provide additional training and implementation details in Section 6 of the Suppl. Material.
\subsection{Results \& Discussion}
\paragraph{The Challenge of Zero-Shot Alignment:} Results in Table~\ref{tab:cross_modal_mechanism} show that standard vision-language foundation models (CLIP~\cite{clip}, BLIP~\cite{li2022blip}, EvaClipv2~\cite{eva02}, SigLIPv2~\cite{tschannen2025siglip}) exhibit poor zero-shot performance, with R@1 scores frequently below 3\%. This underscores a fundamental "semantic-structural gap": generic semantic representations optimized for object-level concepts lack the fine-grained architectural cues essential for localization. 
\begin{table*}[h!]
\centering
\scriptsize
\caption{Cross-modal Vision-Language Retrieval performance. We compare cross-modal foundation models against our language-augmented ({LoRA-MS-}) versions. Bold indicates the best result within each model pair.}
\label{tab:cross_modal_mechanism}
\renewcommand{\arraystretch}{1.5}
\setlength{\tabcolsep}{2pt}
\begin{tabular}{l ccc ccc ccc}
\toprule
\textbf{Model} & \multicolumn{3}{c}{\textbf{Amstertime}} & \multicolumn{3}{c}{\textbf{MSLS-val}} & \multicolumn{3}{c}{\textbf{Pitts30}} \\
\cmidrule(lr){2-4} \cmidrule(lr){5-7} \cmidrule(lr){8-10}
& R@1 & R@5 & R@10 & R@1 & R@5 & R@10 & R@1 & R@5 & R@10 \\
\midrule
CLIP & 2.0 & 7.5 & 12.1 & 2.3 & 6.2 & 10.4 & 10.9 & 29.6 & 43.5 \\
\rowcolor{highlight} LoRA-MS-CLIP & \textbf{11.5} & \textbf{27.5} & \textbf{37.7} & \textbf{38.0} & \textbf{58.4} & \textbf{68.2} & \textbf{49.3} & \textbf{74.4} & \textbf{82.8} \\
\midrule
BLIP & 1.3 & 4.1 & 7.0 & 1.6 & 5.9 & 8.2 & 8.1 & 26.8 & 41.1 \\
\rowcolor{highlight} LoRA-MS-BLIP & \textbf{12.8} & \textbf{31.4} & \textbf{41.1} & \textbf{38.0} & \textbf{60.5} & \textbf{68.4} & \textbf{50.2} & \textbf{74.2} & \textbf{82.8} \\
\midrule
EVA-CLIP-V2 & 2.8 & 8.0 & 12.7 & 3.8 & 9.2 & 13.0 & 11.4 & 30.9 & 44.2 \\
\rowcolor{highlight} LoRA-MS-EVA-V2 & \textbf{11.5} & \textbf{27.9} & \textbf{36.8} & \textbf{32.8} & \textbf{52.8} & \textbf{62.4} & \textbf{41.5} & \textbf{66.8} & \textbf{76.6} \\
\midrule
SigLIP-V2 & 0.9 & 3.1 & 5.5 & 2.6 & 6.5 & 9.7 & 10.1 & 30.1 & 44.3 \\
\rowcolor{highlight} LoRA-MS-SigLIP-V2 & \textbf{13.8} & \textbf{29.0} & \textbf{40.0} & \textbf{35.7} & \textbf{60.5} & \textbf{69.6} & \textbf{49.3} & \textbf{74.6} & \textbf{82.8} \\
\bottomrule
\end{tabular}
\end{table*}
\paragraph{Consistent Efficacy of Parameter-Efficient Fine-Tuning and MS Alignment:} We observe a consistent and substantial performance gain across all evaluated architectures when utilizing Parameter-Efficient Fine-Tuning (PEFT) via LoRA. As detailed in Table~\ref{tab:cross_modal_mechanism}, our LoRA-MS-variants provide an order-of-magnitude improvement over their respective baseline models. For instance, LoRA-MS-SigLIP-V2 reaches an R@1 of 13.8 on Amstertime and 35.7 on MSLS-val, compared to 0.9 and 2.6 for the base model, respectively. Notably, on the Pitts-30k benchmark, LoRA-MS-SigLIP-V2 R@5 performance peaks at 74.6\%, rising from a baseline of 30.0\%. This trend remains stable across different backbones, confirming that LoRA successfully injects domain-specific VPR knowledge while preventing catastrophic forgetting.
\paragraph{Ablation of Training Strategy:} 
Our training strategy ablation (Table~\ref{tab:training_strategies}, rows 2--5) reveals that the synergy between LoRA and the Multi-Similarity (MS) loss is the primary driver of representation convergence. Standard adaptation techniques prove insufficient: both isolated Learned Pooling and full fine-tuning with the MS loss result in near-total failure (R@1 $\approx$ 0.1\%). This collapse occurs because full fine-tuning causes catastrophic forgetting of the foundational VLM's broad semantic priors, which conflict with the rigid visual-discriminative objectives of the VPR task. 

By contrast, applying LoRA jointly to both the vision and text towers restricts adaptation to low-rank updates, allowing each encoder to align its representations toward place-discriminative cues while preserving its pre-trained priors. Concurrently, the MS loss provides the hard-pair mining necessary to separate geographically similar locations, a signal that standard contrastive objectives (such as the original BLIP loss) lack, leading to markedly lower performance. Ultimately, coupling the parameter-efficient capacity of LoRA with the hard-mining capabilities of the MS loss is what enables the framework to successfully bridge the semantic-structural gap.
\begin{table}[h!]
\centering
\scriptsize
\caption{Comparison of training strategies for Cross-modal VPR over BLIP model.}
\label{tab:training_strategies}
\setlength{\tabcolsep}{3.8pt}
\renewcommand{\arraystretch}{1.2}
\begin{tabular}{ll cccc}
\toprule
\textbf{Training Strategy} & \textbf{Loss} & \multicolumn{4}{c}{\textbf{Amstertime (Recall@K)}} \\
\cmidrule(lr){3-6}
& & R@1 & R@5 & R@10 & R@20 \\
\midrule
Zero-shot & Frozen Encoders & 1.3 & 4.1 & 7.0 & 12.3 \\
LLP & MS & 0.1 & 0.5 & 1.0 & 2.4 \\
Full Fine-tuning & MS & 0.1 & 0.4 & 0.9 & 1.6 \\
LoRA (All Linear, $r=64$) & Contrastive & 0.1 & 0.4 & 0.8 & 1.6\\ 
\midrule
\rowcolor{gray!10} LoRA (All Linear, $r=64$) & \textbf{MS} & \textbf{12.8} & \textbf{31.4} & \textbf{41.1} & \textbf{53.2} \\
\bottomrule
\end{tabular}
\end{table}

\paragraph{LoRA Configuration and Coverage:} The ablation results in Table~\ref{tab:lora_ablation} indicate that the scope of adaptation is as critical as the rank itself. Targeting all linear layers outperforms targeting only $Q, K, V$ matrices. Extending coverage across all projections with rank $r=64$ provides the necessary capacity to capture complex environmental semantics, yielding a peak R@1 of 12.8\% on Amstertime. This suggests that the structural nuances required for VPR are distributed throughout the Transformer blocks, necessitating a comprehensive adaptation of both attention and feed-forward mechanisms.
\begin{table}[h]
\centering
\scriptsize
\caption{Ablation of LoRA hyperparameters and Loss functions over BLIP model. }
\label{tab:lora_ablation}
\setlength{\tabcolsep}{3.8pt}
\renewcommand{\arraystretch}{1.2}
\begin{tabular}{lc l cccc}
\toprule
\textbf{Target Layers} & \textbf{Rank ($r$)} & \textbf{Loss} & \multicolumn{4}{c}{\textbf{Amstertime (Recall@K)}} \\
\cmidrule(lr){4-7}
& & & R@1 & R@5 & R@10 & R@20 \\
\midrule
LoRA (QKV) & 16 & Contrastive & 4.0 & 10.1 & 13.9 & 19.9 \\
LoRA (QKV) & 16 & MS & 8.9 & 25.0 & 34.0 & 44.8 \\
LoRA (QKV) & 64 & MS & 10.6 & 29.1 & 38.0 & 49.3 \\
LoRA (All Linear) & 64 & Contrastive & 0.1 & 0.4 & 0.8 & 1.6 \\
\midrule
\rowcolor{gray!10} \textbf{LoRA (All Linear)} & \textbf{64} & \textbf{MS} & \textbf{12.8} & \textbf{31.4} & \textbf{41.1} & \textbf{53.2} \\
\bottomrule
\end{tabular}
\end{table}

\paragraph{Impact of Descriptive Density:} Table~\ref{tab:rerank_crossmodal_short_text}, LoRA-MS-BLIP's performance scales directly with narrative length, and consistently outperforms the zero-shot BLIP baseline. Specifically, increasing the average query length from 15 to 60 words yields a 100\% relative R@1 gain for LoRA-MS-BLIP (6.5\% to 13.2\%). In contrast, the zero-shot baseline peaks at 35 words ($R@1=3.2\%$) before collapsing to 1.3\% as complexity increases. This divergence suggests that while standard foundation models are optimized for short, object-centric captions, our training paradigm enables the model to resolve complex scenes by aggregating fine-grained structural cues distributed throughout a dense, multi-element narrative.
\begin{table}[h!]
  \centering
  \caption{Impact of Query Length on Cross-Modal Retrieval. Descriptive statistics are reported in word counts.} 
  \label{tab:rerank_crossmodal_short_text}
  \scriptsize 
  \renewcommand{\arraystretch}{1.2}
  \setlength{\tabcolsep}{3.5pt}
  \begin{tabular}{l ccc cccc}
  \toprule
  \textbf{Model} & \multicolumn{3}{c}{\textbf{Query Length Statistics}} & \multicolumn{4}{c}{\textbf{Amstertime (Recall@K)}} \\
  \cmidrule(lr){2-4} \cmidrule(lr){5-8}
  & Min & Avg $\pm$ Std & Max & R@1 & R@5 & R@10 & R@20 \\
  \midrule
  \rowcolor{gray!10} \multicolumn{8}{l}{\textit{Full Narratives}} \\
  BLIP    & 6  & 15 $\pm$ 1.9 & 20 & 2.8 & 8.8 & 13.2 & 19.3 \\
  LoRA-MS-BLIP & 6  & 15 $\pm$ 1.9 & 20 & 6.5 & 17.1 & 24.9 & 34.8 \\  \midrule  
  BLIP    & 10 & 25 $\pm$ 4.9 & 44 & 3.1 & 9.7 & 13.6 & 18.6 \\
  LoRA-MS-BLIP & 10 & 25 $\pm$ 4.9 & 44 &  9.0 & 24.7 & 34.0 &42.1 \\
  \midrule
  BLIP    & 11 & 35 $\pm$ 7.2 & 74 &  3.2 & 7.7& 13 &	19.0 \\
  LoRA-MS-BLIP & 11 & 35 $\pm$ 7.2 & 74 &  11.0 & 25.9 &35.8 & 45.9 \\    
  \midrule
  BLIP    & 10 & 60 $\pm$ 22 & 184 & 1.3 & 4.1 & 7.0 & 12.3 \\
  LoRA-MS-BLIP & 10 & 60 $\pm$ 22 & 184 & 13.2 & 31.8 & 42.3 & 52.8 \\    
  \bottomrule
  \end{tabular}
\end{table}


\section{Limitations and Future Work} \label{sec:limitations}
Our framework exhibits a few limitations that highlight promising directions for future research. First, we observe diminishing returns when applying language augmentation to state-of-the-art self-supervised backbones; when underlying visual features are already highly discriminative, the introduced linguistic context can occasionally inject sub-optimal priors. Second, despite the substantial performance gains enabled by our LoRA-MS paradigm, a significant performance gap persists between cross-modal and uni-modal retrieval. While our baseline represents a major leap over zero-shot VLM performance, ``blind'' localization from text descriptions alone remains markedly less reliable than image-based retrieval. We explicitly highlight closing this cross-modal gap as an open challenge and a call to action for the community. 

Finally, to isolate and evaluate the immediate utility of language augmentation, this study prioritized late-fusion architectures. Although our preliminary experiments with mid-fusion configurations resulted in performance degradation, the trade-off between enforcing tight semantic alignment via early-fusion strategies and preserving fine-grained, discriminative visual features remains a complex architectural dynamic. A systematic evaluation of these early vs. late-fusion trade-offs is beyond the scope of this work. Consequently, future research will investigate more sophisticated cross-modal alignment objectives, advanced multi-stage fusion mechanisms, and the integration of additional modalities to bridge these remaining performance gaps.
\section{Conclusion} \label{sec:conclusion}
This work introduces LaVPR, a large-scale benchmark that extends vision-only datasets with high-density linguistic descriptions to investigate multi-modal and cross-modal localization. Our analysis demonstrates that language serves as a critical robustness anchor; specifically, our ADS-LLP mechanism yields consistent gains and superior computational efficiency compared to the vertical scaling of visual transformers. Finally, we establish that the synergy between LoRA and Multi-Similarity (MS) loss is essential for effective cross-modal retrieval, providing order-of-magnitude improvements over standard baselines.

\clearpage
\setcounter{table}{0}
\setcounter{section}{0}
\section*{Supplementary Material}
The following sections provide a deep dive into the experimental and algorithmic foundations of our work:
\begin{itemize}
    \item Section~\ref{sec:appendix_generation} details \textit{LaVPR generation} and its specialized subsets, including the standardized prompts used for large-scale description generation, curation protocols like signage-based filtering for Amstertime-La, and synthetic augmentations for MSLS-Blur/Weather. It further illustrates the impact of utilizing different Vision Large Language Models (VLLMs).
    \item Section~\ref{sec:appendix_curation} delineates the \textit{curation pipeline}, presenting a formal algorithmic walkthrough of our verification process alongside an ablation study of our cleaning strategy.
    \item Section~\ref{subsec:appendix_extended_mm} provides extended results for \textit{multi-modal fusion} via concatenation (CAT) and examines the performance impact of varying the textual encoder architecture.
    \item Section~\ref{sec:appendix_complexity} presents an \textit{efficiency and complexity analysis}, complementing Table 7 in the main text by providing explicit GFLOP and parameter count derivations to assess the trade-offs between architectural scaling and retrieval performance. 
    \item Section~\ref{subsec:appendix_multimodal_impl} outlines comprehensive \textit{training and implementation} specifications for the \textit{multi-modal fusion} framework.
    \item Section~\ref{subsec:appendix_crossmodal_impl} documents the \textit{implementation and training} details of the \textit{cross-modal retrieval} paradigm. 
    \item Section~\ref{subsec:appendix_qualitative_results} showcases \textit{qualitative examples from the LaVPR dataset} and provides visual comparisons between the baseline MixVPR and our proposed La-MixVPR to illustrate success cases in challenging urban environments.
\end{itemize}
\section{LaVPR Generation: Extended Details}
\label{sec:appendix_generation}
We provide selected samples from the LaVPR dataset in Figures 1-4. 
\subsection{Primary Generation Prompt}\label{subsubsec:appendx_generation_prompt}
To ensure consistent, high-density descriptions across 651,865 image-text pairs, we utilized Gemini 2.5 Flash with the following standardized prompt:

\begin{quote} "Describe this location for visual place recognition. Focus on: 1) Scene type and setting, 2) Distinctive landmarks and architecture, 3) Unique visual patterns/colors/textures, 4) Spatial layout, 5) Key identifying features that distinguish this place from similar locations. Be specific about permanent visual elements, avoid temporary objects like people, cars, weather and lighting conditions, provide textual descriptions of items you are certain about only. the output is one line of text listing the items from left to right, separated by commas." \end{quote}

We experimented with similar prompts and found the VLLM to yield relatively stable descriptions. While we opted for a high performing VLLM (Gemini 2.5 Flash) for LaVPR, Fig.~\ref{fig:appendix_generation} illustrates that even a weaker model such as InternCL-38B can generate high quality captions.
\begin{figure}[htbp] 
    \centering
    \includegraphics[width=0.8\linewidth]{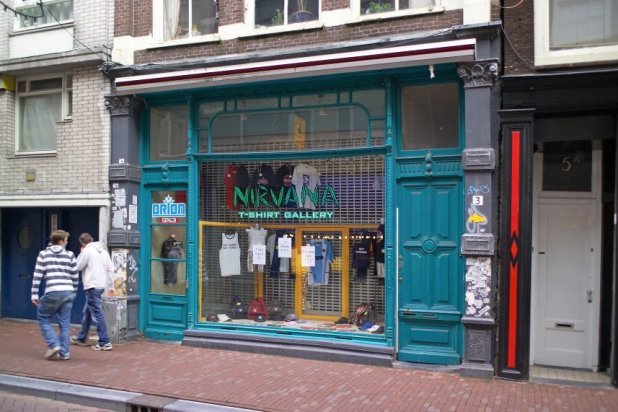}
    \captionsetup{font=scriptsize} 
    \caption{Example image from the Amstertime subset. \textit{Generated description in LaVPR:} Light-colored brick building with multiple windows, blue ground-level door with a round window, dark gray ornate pillar, teal/turquoise framed glass door with "ORION" logo, teal/turquoise shopfront with display windows behind a metal security gate, green and purple "NIRVANA T-SHIRT GALLERY" sign, red and white striped awning, teal/turquoise ornate double door, dark gray ornate pillar with number "3" and graffiti, black facade with vertical red and black diamond pattern, white door with number "54" above it, brick sidewalk.}
    \label{fig:amstertime_example_1}
\end{figure}\textbf{}

\begin{figure}[htbp] 
    \centering
    \includegraphics[width=0.8\linewidth]{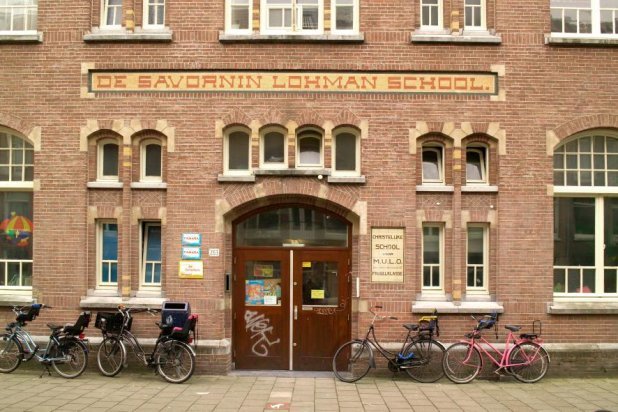}       
    \captionsetup{font=scriptsize} 
    \caption{ Example image from the Amstertime subset. \textit{Generated description in LaVPR:} A multi-story reddish-brown brick building facade with numerous white-framed windows and light-colored stone accents; from left to right: a partial large arched window with white grilles, a section of brick with a vertical stack of two rectangular white-framed windows, a large rectangular sign integrated into the brick facade reading "DE SAVORNIN LOHMAN SCHOOL." with orange lettering and light-colored borders, a section of brick with two upper small arched windows and two lower rectangular white-framed windows, a central large arched brick entrance containing double dark wooden doors, above which are three small arched windows, a rectangular light-colored plaque on the brick wall inscribed with "CHRISTELIJKE SCHOOL M.U.L.O. TEGELAAR", a section of brick with a vertical stack of two rectangular white-framed windows, and a partial large arched window with white grilles.}
\end{figure}

\begin{figure}[htbp] 
    \centering
    \includegraphics[width=0.8\linewidth]{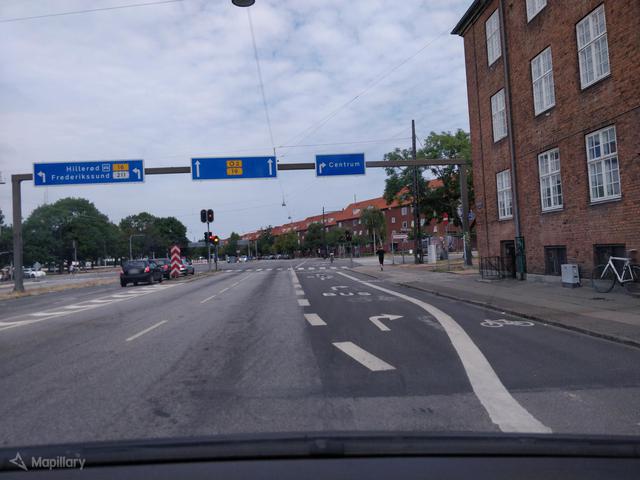}
    \captionsetup{font=scriptsize} 
    \caption{ Example image from the MSLS-Val subset. \textit{Generated description in LaVPR:} Green trees, blue overhead road sign pointing left to "Hillerød 16" and "Frederikssund 211", asphalt road with white lane markings, red traffic lights at an intersection, blue overhead road sign pointing straight to "O2 16", blue overhead road sign pointing straight to "Centrum", asphalt road with "BUS" painted on it and a white left-turn arrow marking a bicycle lane, rows of red-brick buildings with white-framed windows, large red-brick building with numerous white-framed windows, a sidewalk.}
\end{figure}\textbf{}

\begin{figure}[htbp] 
    \centering
    \includegraphics[width=0.8\linewidth]{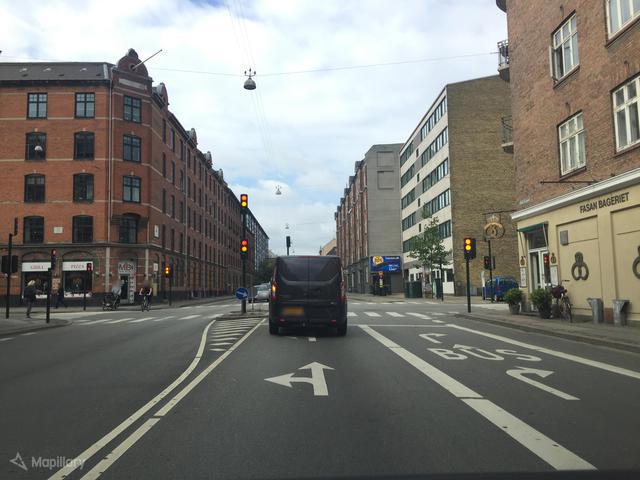}
    \captionsetup{font=scriptsize} 
    \caption{ Example image from the MSLS-val subset. \textit{Generated description in LaVPR:} Large multi-story red brick corner building with dark windows, shops on ground floor; pedestrian crossing markings on asphalt road; double white lines on road separating lanes; white left-turn arrow painted on road; traffic light pole; sequence of buildings further down the street including modern light-colored building with horizontal windows; blue circular street sign; multi-story red brick building on right; light yellow-beige shop facade with "FASAN BAGERIET" sign and golden pretzel symbols; "BUS" lane marking painted on road.}
\end{figure}

\begin{figure}[h!]
    \centering
    \includegraphics[width=0.9\linewidth]{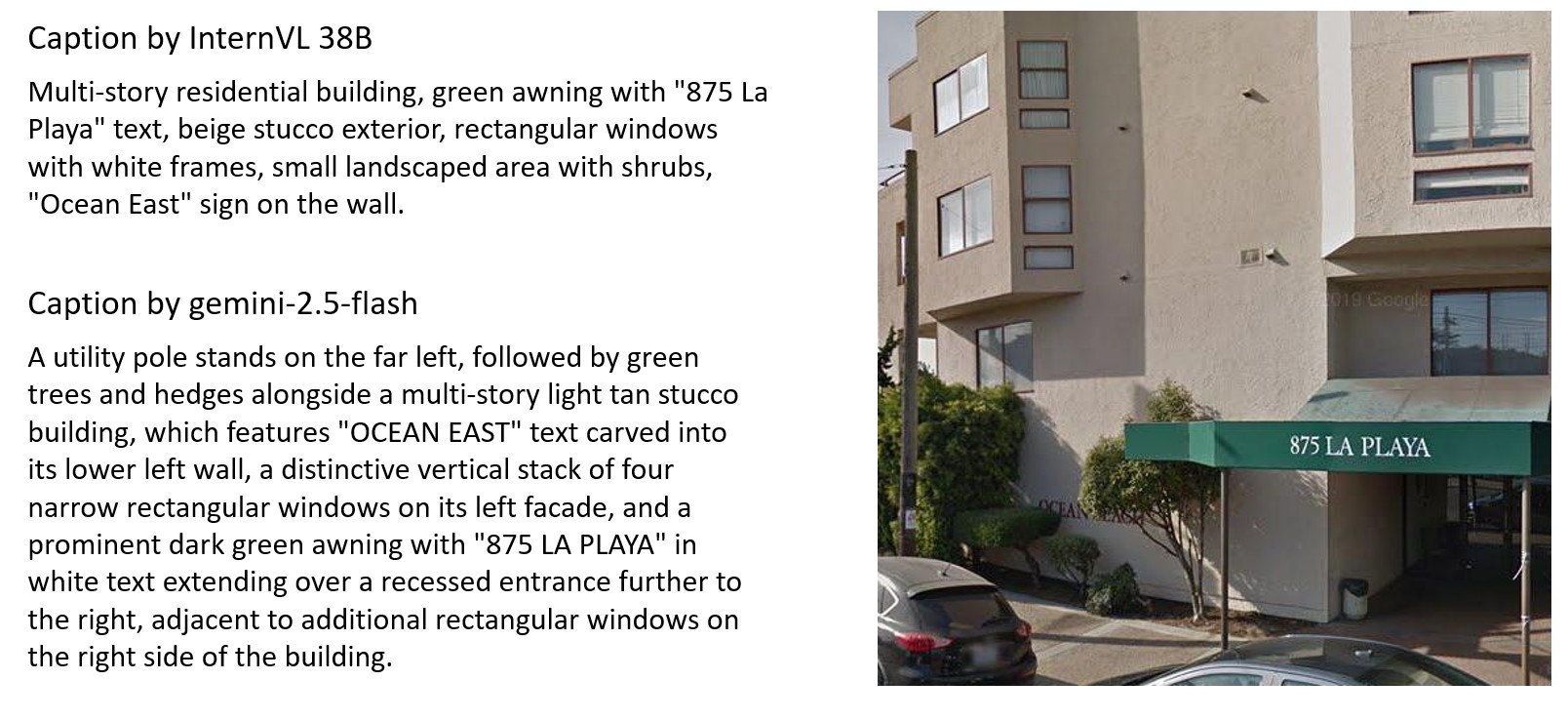}
    \captionsetup{font=scriptsize} 
    \caption{ Comparison of caption generated by open VLM InternCL-38B vs. Gemini-Flash-2.5. Both captions shows high quality descriptions. }
    \label{fig:appendix_generation}
\end{figure}

\subsection{Specialized Dataset Generation 
Protocols}\label{subsubsec:appendix_spec_datasets}
To evaluate the limits of language-vision complementarity, we curated specialized subsets targeting specific VPR challenges: semantic distinctiveness (Amstertime-La) and visual degradation (MSLS-Blur/Weather). These subsets facilitate a controlled analysis of how linguistic context compensates for pixel-level information loss.

\begin{figure}[h!]
    \centering
    \includegraphics[scale=0.5]{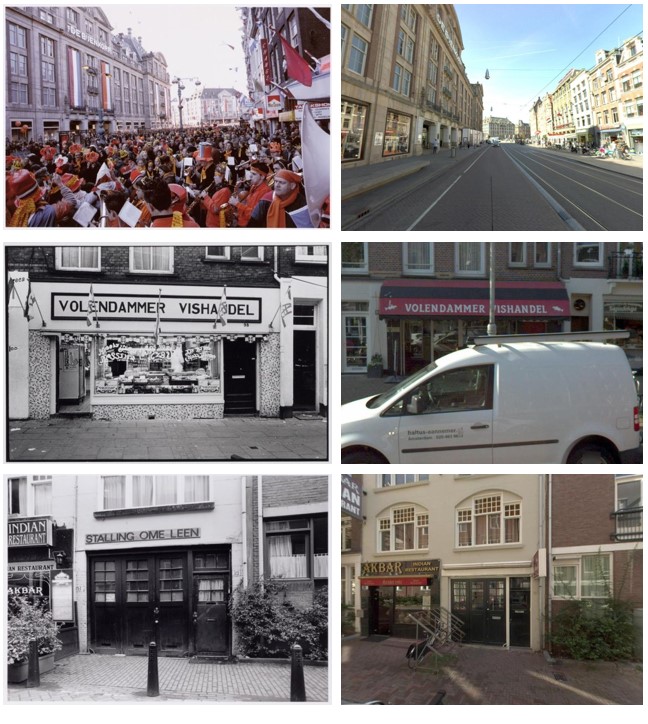}
    \caption{Example pairs from the Amstertime-La dataset. The dataset is curated by identifying shared signage and scene text between query (left) and database (right) images from the Amstertime dataset, to ensure fine-grained semantic alignment.}
    \label{fig:your_amstertime_la_example}
\end{figure}

\paragraph{Signage-based Filtering (Amstertime-La):} This subset targets locations with high discriminative potential. We utilized Gemini 2.5 Flash to identify and filter database-query pairs where specific signage text matched exactly across both images. Qualitative examples of these high-precision alignments are shown in Figure~\ref{fig:your_amstertime_la_example}.
We used the following prompt for filtering: \begin{quote} "filter images of same place where the sign text is the same between database and query". 

for example: 

queries/image1.jpg, "INDIAN RESTAURANT AKBAR signage, framed menu display, street number 151, planter with green foliage, dark bollard, brick-patterned pavement, STALLING OME LEEN sign, large dark double doors with glass panes, smaller dark single door with glass panes and mail slot, street number 19, dark bollard, brick-patterned pavement, brick wall with dense green ivy, ground-floor window with white curtains."

database/image1.jpg, "Leftmost building fragment with partial sign, light-colored building with two upper arched multi-paned windows, AKBAR INDIAN RESTAURANT sign, red awning, restaurant entrance door, dark double-door entrance, metal bike rack, brown brick building with upper window and black balcony railing, lower white-framed window, green shrubbery, reddish-brown brick pavement."

both have sign with text of AKBAR INDIAN RESTAURANT.
put the result in table format." \end{quote}

\begin{figure}[h!]
    \centering
    \includegraphics[scale=0.35]{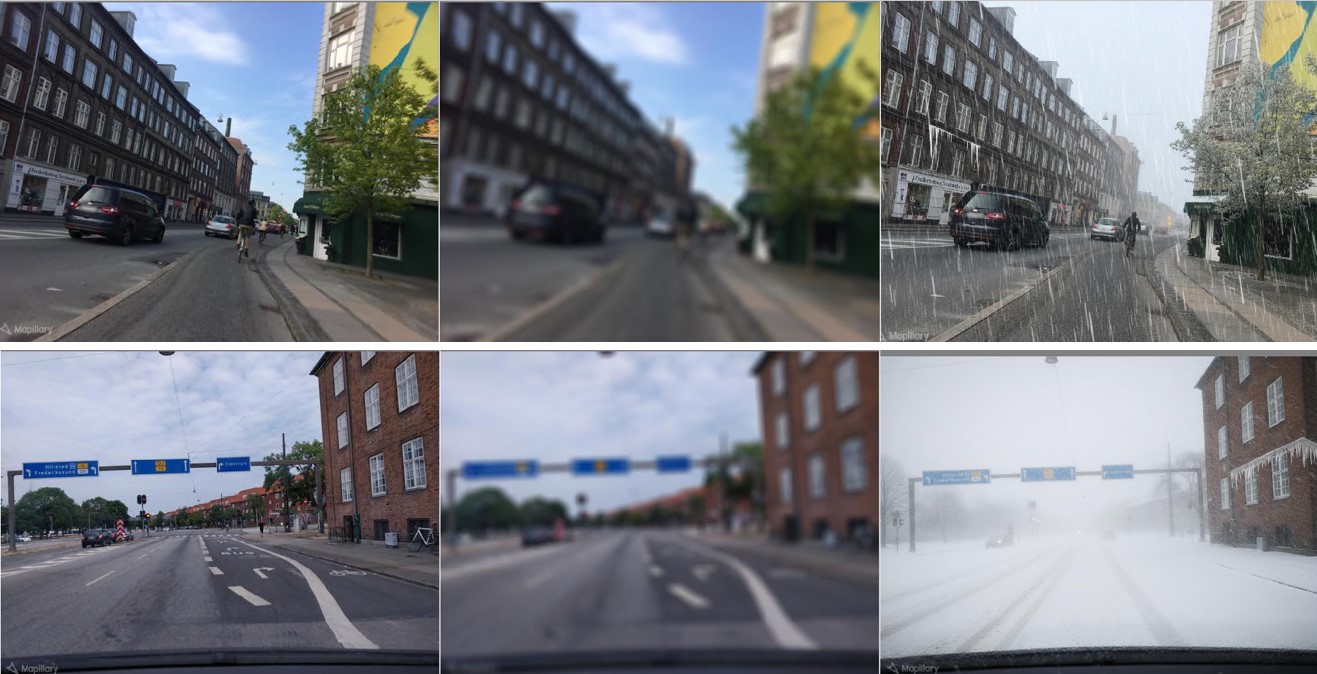}
    \caption{Example images from MSLS-Blur and MSLS-Weather subset, left is original query image, middle is blurred augmentation, right is augmented with rain, snow and fog. }
    \label{fig:msls_blur_example}
\end{figure}

\paragraph{Visual Degradation Synthesis (MSLS Subsets):} To further simulate extreme operational environments, we applied synthetic transformations to the MSLS-val queries. 
For BLUR augmentations, we used Photoshop's Lens Blur tool. We further utilized gemini-nano-banana to generate adverse weather conditions (e.g., heavy rain, snow, and fog) with the following prompt:
\begin{quote}
    "Transform this image. Make it look like snow, with strong rain and a lot of fog. Add icicles to the buildings and trees."
\end{quote}
The visual characteristics of these augmentations are illustrated in Figure~\ref{fig:msls_blur_example}.

\section{LaVPR Curation: Extended Details}
\label{sec:appendix_curation}
\subsection{Curation Pipeline: Algorithmic Walkthrough}\label{subsec:appendix_pipeline_algo}
Enriching standard VPR training and evaluation datasets with aligned natural language descriptions, using a state-of-the-art Vision–Language Model, requires a strategy that accounts for model reliability constraints. Although Gemini Flash 2.5 is widely regarded as a state-of-the-art model, we lacked a concrete quantitative assessment of its performance, beyond general expectations reported in the literature \cite{google2025gemini}.

Given the natural variability in human image descriptions, we prioritized descriptive accuracy over completeness. Consequently, we designed and developed a pipeline to validate the presence of objects and scene elements mentioned in the generated descriptions.

\begin{figure}[t]
    \centering
    \includegraphics[scale=0.22]{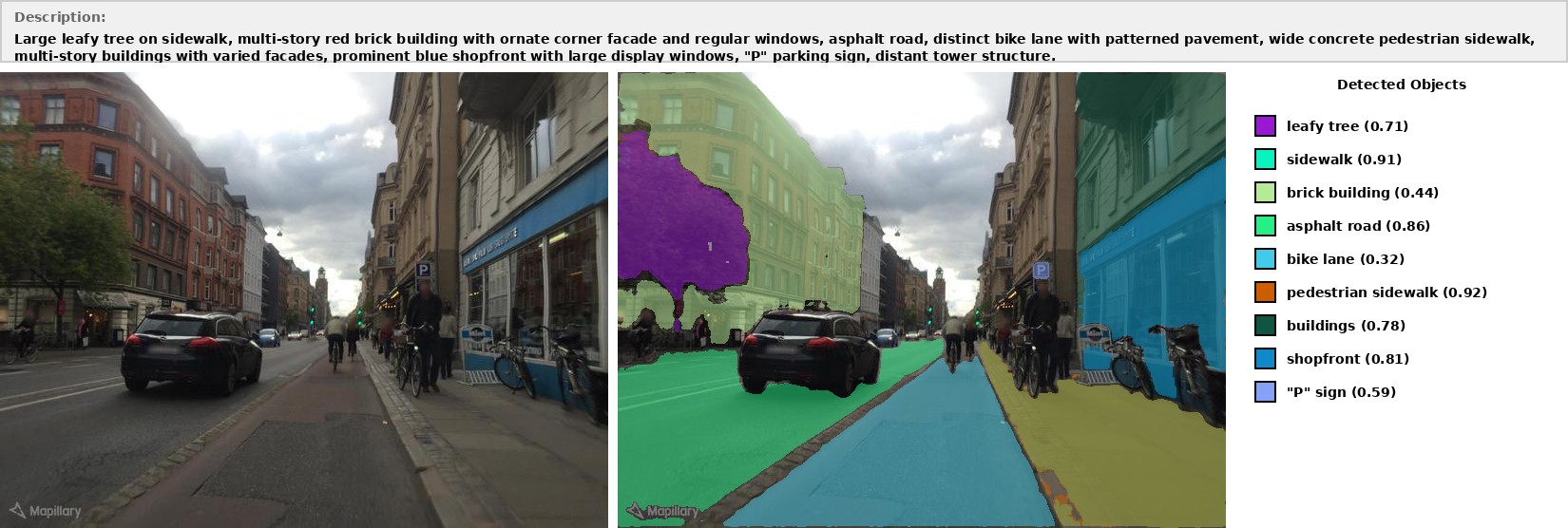}
    \caption{Qualitative visualization of Step 2: Open-set object detection using SAM3. 
    Given an input image and a set of object queries extracted from the generated description, SAM3 produces segmentation masks for visually identifiable objects.}
    \label{fig:sam3_example}
\end{figure}

Given an image–description pair $(I, D)$, our pipeline applies the following four steps:
\begin{enumerate}
    \item \textbf{Object Extraction.}  
    Given the generated description $D$, we use a large language model (LLM) to extract a list of mentioned objects $\mathcal{O}_D$.
    \item \textbf{Open-Set Object Detection.}  
    We provide the image $I$ together with $\mathcal{O}_D$ as input to an open-set object detector, to localize all mentioned objects in the image. Figure~\ref{fig:sam3_example} provides an example output of this step.
    \item \textbf{VLM-Based Verification.}  
    For each object that is not detected in the previous step, we perform an additional verification step using a vision–language model. 
    Specifically, we ask a binary (yes/no) question to determine whether the object can be visually identified in the image.
    \item \textbf{Textual Object Filtering.}  
    From the remaining objects that are still not found in the image, we use an LLM to automatically filter out objects categories that correspond to textual elements, contain special characters, or objects that are deemed undetectable (for example: "sunlight").
\end{enumerate}
After applying this procedure to all image–description pairs, we aggregate the remaining objects across pairs, group them by unique object notion, and rank the groups by frequency. Next, we manually inspect the top object groups (we set $K=20$ for the test and validation datasets) and assign each group to one of the following categories (qualitative illustrations presented in Figure~\ref{fig:object_failure_examples}):
\begin{enumerate}
    \item \textbf{Hallucination:} Objects that do not appear in the image and are considered hallucinated by the Vision–Language model we used to generate the text descriptions.
    \item \textbf{Visually Degraded:} Objects that are extremely small, heavily occluded, or located at the image boundaries.
    \item \textbf{Out-of-Distribution (OOD):} Long-tail or rare objects that fall outside the models' distribution used in steps 2 and 3 of our pipeline.
\end{enumerate}
For each group, the labeling decision is made by visually inspecting a small sample of object instances from that group. Finally, to mitigate hallucinations, we edit the generated descriptions by removing only the objects belonging to the Hallucination category from their generated descriptions they were extracted from.
\subsection{Curation Analysis}

\begin{figure}[t]
    \centering
    \includegraphics[scale=0.1]{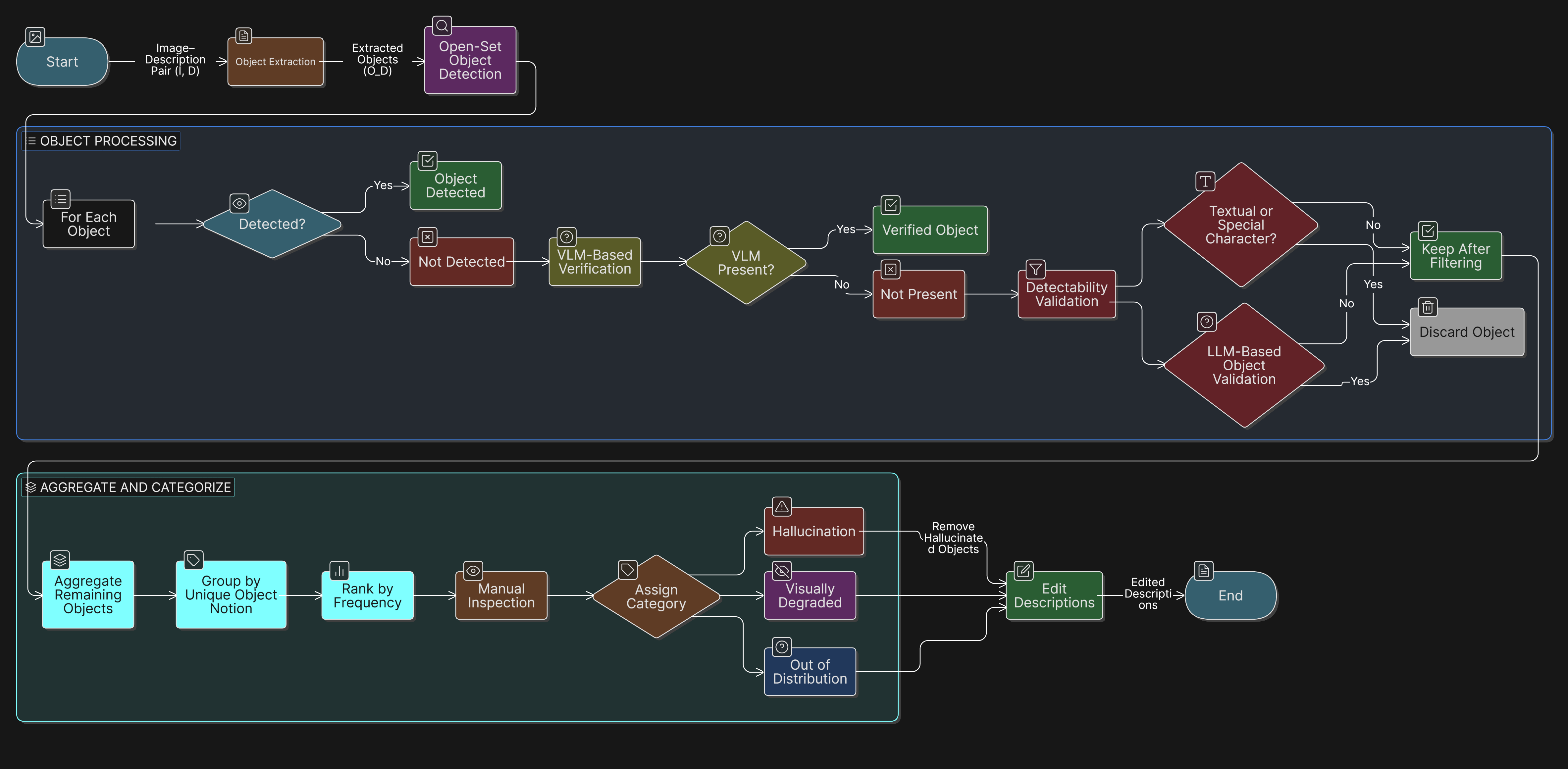}
    \caption{LaVPR Curation Pipeline Diagram.}
\label{fig:pipeline_verfication_arch}
\end{figure}
\begin{figure}[h]
    \centering
    \includegraphics[width=0.32\linewidth]{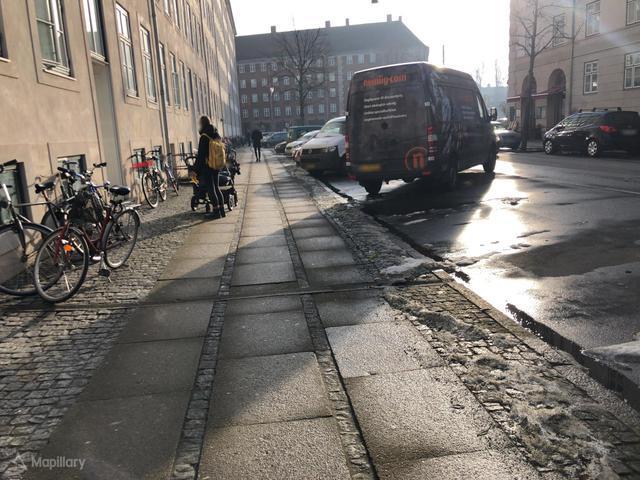}
    \hfill
    \includegraphics[width=0.32\linewidth]{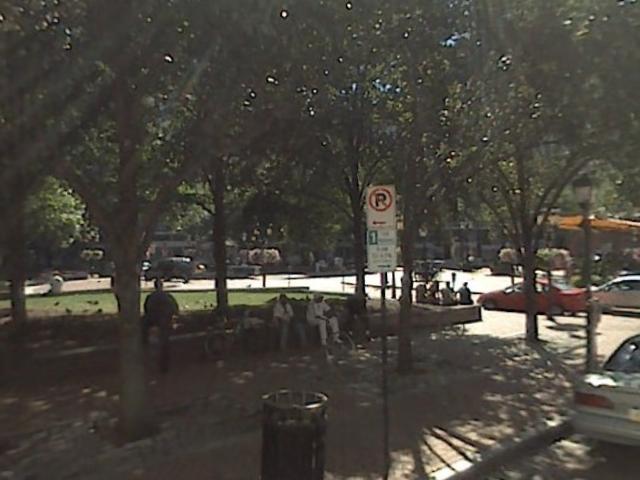}
    \hfill
    \includegraphics[width=0.32\linewidth]{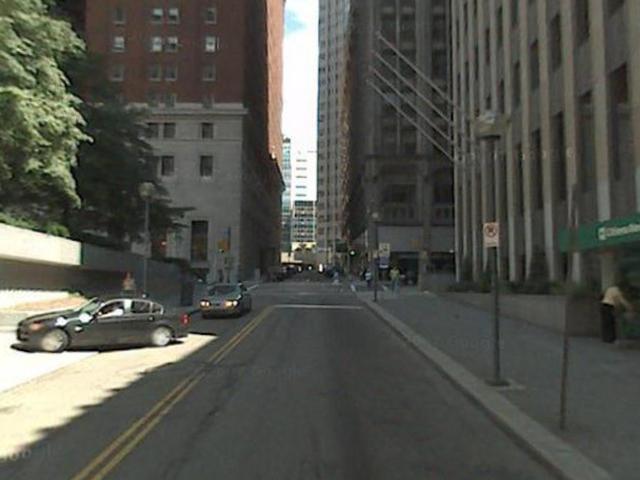}
    \caption{Examples of unresolved object categories.
    \textbf{Left:} Hallucinated object (low dark metal railing).
    \textbf{Middle:} Visually degraded object (buildings heavily obscured).
    \textbf{Right:} Out-of-distribution object (embankment).}
    \label{fig:object_failure_examples}
\end{figure}

\begin{table}[t]
\centering
\scriptsize
\caption{Unified Pipeline Statistics: Models and Filtering Progression. ``Remaining'' columns indicate the \% of descriptions that passed all checks up to that step.}
\label{tab:pipeline_unified}
\renewcommand{\arraystretch}{1.3}
\begin{tabular}{lccccccc}
\toprule
\textbf{Pipeline Stage} & \textbf{Model} & \textbf{Primary Function} & \multicolumn{5}{c}{\textbf{Descriptions Remaining (\%)}} \\
\cmidrule(lr){4-8}
& & & \textbf{Ams.} & \textbf{MSLS-v} & \textbf{MSLS-c} & \textbf{P-T} & \textbf{P-V} \\
\midrule
1. Entity Extraction & Phi-3.5-mini & Entity extraction & 100 & 100 & 100 & 100 & 100 \\
2. Grounding & SAM3 & Spatial localization ($\tau=0.2$) & 68 & 66 & 66 & 67 & 65 \\
3. Verification & Qwen2-VL-7B & Binary hallucination check & 39 & 36 & 35 & 33 & 32 \\
4. Object Filtering & Phi-3.5-mini & Selective token removal & 29 & 28 & 26 & 23 & 23 \\
\bottomrule
\end{tabular}
\end{table}
{We ran our pipeline on the validation and test datasets listed in Table 1 in the main manuscript}.
Across all datasets, approximately 26\% of the combined database and query samples remained after Step 4 (see Table ~\ref{tab:pipeline_unified}). 
After aggregation, we obtained 6,336 unique object groups. Manual inspection of the 20 largest groups showed that 15 corresponded to out-of-distribution (OOD) objects, while the remaining 5 contained visually degraded content. Overall, descriptions flagged as containing hallucinations and subsequently sent for cleaning accounted for $\sim1\%$ of the inspected images within the top-20 categories.
Figure~\ref{fig:pipeline_verfication_arch} further illustrates the entire pipeline schema. 
\subsection{Manual Validation of In-Scene Signage}
We perform further human-in-the-loop verification on Amstertime-La and MSLS-Val to ensure the accuracy of in-scene signage. This step specifically addresses textual elements that elude automated segmentation, providing a high-precision ground truth for discriminative scene text. We manually analyzed 117 images and their corresponding texts. The analysis demonstrated that descriptions are predominantly accurate with rich detail that supports fine-grained location discrimination, as illustrated in Figure~\ref{fig:amstertime_example_1}. Nevertheless, minor inaccuracies were observed, primarily in signage labels that were occluded or due to occasional character recognition errors, still most of the signage labels were correct, reaching an accuracy of 93.2\% on our samples images
\subsection{Cleaning Strategy Ablation}
\begin{table}[h]
  \centering
  \scriptsize
  \caption{Effect of cleaning strategies on retrieval performance. Results are reported for {La-MixVPR} using {CAT fusion} and BGE-L as the text encoder.} 
\renewcommand{\arraystretch}{1.2}
  \begin{tabular}{@{}l cccc@{}}
    \toprule
    \textbf{Cleaning Strategy} & \multicolumn{4}{c}{\textbf{Amstertime (Recall@K)}} \\
    \cmidrule(l){2-5}
    & R@1 & R@5 & R@10 & R@20 \\
    \midrule
    None (Baseline) & \textbf{37.0} & \textbf{58.2} & \textbf{65.6} & \textbf{71.6} \\
    Hybrid          & \textbf{37.0} & \textbf{58.2} & 65.5          & \textbf{71.6} \\
    Full            & 36.6          & 58.0          & 65.0          & \textbf{71.6} \\
    \bottomrule
  \end{tabular}
    \label{tab:clean_strategy}
\end{table}
We evaluate several dataset refinement strategies to identify the optimal balance between descriptive density and semantic accuracy (Table~\ref{tab:clean_strategy}). Under the 'Full Clean' strategy, an LLM is utilized to purge all objects identified by our pipeline as candidate hallucinations. Conversely, the 'Hybrid' strategy, detailed in Section~\ref{subsec:appendix_pipeline_algo}, employs manual verification of high-frequency flagged categories to selectively determine which entities to remove. For both methodologies, we utilize Llama-3.3-70B-Instruct for automated entity removal. We also compare these strategies to a permissive strategy, which applies description curation. The marginal performance variance observed across these strategies suggests that noise level is initially relatively low (consistent with our curation analysis). Results also indicate a slight advantage to permissive strategies, suggesting that models can handle local sparse hallucinations (one hallucinated object within a rich and dense description).
\subsection{Pipeline Implementation: Prompts and Models}
\begin{itemize}
\item \textbf{\textit{Step 1 (caption $\rightarrow$ objects + stuff extraction,
\href{https://huggingface.co/microsoft/Phi-3.5-mini-instruct}{Phi-3.5-mini-instruct}):}}
\begin{tcolorbox}[
  breakable,
  colback=gray!3,
  colframe=black!70,
  boxrule=0.8pt,
  arc=2mm,
  left=4pt,
  right=4pt,
  top=4pt,
  bottom=4pt,
  fontupper=\ttfamily
]

    I want to use both an object detector and a segmentation model to check
    the correctness of an image caption obtained by an image caption model.
    Can you help to parse the caption below and list: (1) All objects that could be detected with an object detection model in the image, and (2) All uncountable regions (stuff) that could be segmented in the image. Examples of stuff: road, grass, pavement, wall, sidewalk, water, ground. Stuff are materials, surfaces, or regions rather than individual objects. Please ignore sky. Please ignore dynamic objects like people and cars Please only list the object/region names and ignore descriptions like colors. Please use singular for all listed items. Please do not list signs or surfaces with written words. Here are some examples of the desired behavior in the format of caption and expected output:\\
    Example 1 \newline
    Caption: Green utility box/door set in a brick wall, dense green foliage and trees along the left side, paved sidewalk, asphalt road with double yellow lines and white ""DISABLED"" marking, paved sidewalk, black ornate street lamp post, dark metal railings, row of brown brick terraced houses with white window frames and white ground floor sections, distant buildings.
    \newline
    utility box. brick wall. green foliage. tree. sidewalk. asphalt road. street lamp post. railings.  terraced houses.  white window frames.  buildings. \#\#\# END\_OF\_LIST \#\#\#
    \newline
    Example 2
    \newline
    Caption: White painted brick wall (upper left), dark brick facade with white-framed multi-pane sash window (upper left), dark brick facade with black-framed door and black iron balcony (upper center), dark brick facade with white-framed multi-pane sash window (upper right), dark grey tiled sloped roof (lower center), black framed door/window unit (lower left), large white-framed multi-pane sash window (lower center), white wall with black wall-mounted lantern and silver intercom/doorbell panel (lower right), black garage door with a narrow window above it (lower right), vertical cream/yellow pipe (far right).
    \newline
    wall. window. door. balcony. roof. pipe \#\#\# END\_OF\_LIST \#\#\#
    \newline
    Example 3
    \newline
    Caption: River Thames with distant buildings and boats, ornate dark lamppost with criss-cross patterns, bridge parapet with dark top rail and decorative light-on-red criss-cross panels, blue road sign with bus and bicycle pictograms, light-colored lamppost on a column base, grey paved walkway, ""BUS"" road marking with a red line.
    \newline
    River. lamppost. bridge parapet. road. sign. walkway  \#\#\# END\_OF\_LIST \#\#\#
    \newline
    Now do the same for this new caption.
    Caption: {caption}.
    Please concatenate all objects and stuff together with "." as separation.
    Do not add any explanation, notes, headings, or markdown; output only the
    concatenated object and region names.
    After the last item, output exactly \#\#\# END\_OF\_LIST \#\#\# and nothing else.
\end{tcolorbox}
  \item \textbf{\textit{Step 3 (vision--language yes/no checker,
  \href{https://huggingface.co/Qwen/Qwen2-VL-7B-Instruct}{Qwen2-VL-7B-Instruct}):}}
\begin{tcolorbox}[
  breakable,
  colback=gray!3,
  colframe=black!70,
  boxrule=0.8pt,
  arc=2mm,
  left=4pt,
  right=4pt,
  top=4pt,
  bottom=4pt,
  fontupper=\ttfamily
]
Answer strictly with 'yes' or 'no'. Is there a '{obj}' clearly visible in this image?
\newline
Example:
\newline
Input: Answer strictly with 'yes' or 'no'.
\newline
Is there a 'tree canopies' clearly visible in this image?
\newline
Expected output: yes.
\end{tcolorbox}
\end{itemize}

\section{Extended Results for Multi-Modal Place Recognition} \label{subsec:appendix_extended_mm}
Table~\ref{tab:lavpr_cat_multimodal} shows the results of language-augmented (La-) VPR models with CAT fusion. Despite its simplicity, concatenation yields substantial gains, most notably for smaller CNN-based backbones.

Table~\ref{tab:effect_of_text_encoder} shows the multi-modal results for different textual encoder backbone. Larger state-of-the-art language encoders (e.g., ModernBERT-Large~\cite{warner2025smarter}) excel in semantically dense environments like Amstertime-La (61.3\% R@1). However, in degraded visual conditions (Blur/Weather), performance is uniform across encoder sizes (e.g., BGE-S and BGE-L). This suggests that while complex scenes require sophisticated linguistic discrimination, simple "semantic anchoring" is sufficient to overcome visual noise.
\begin{table*}[h]
\centering
\scriptsize
\caption{Comparison of vision-only VPR methods vs. our Language-Augmented (La-) variants using {CAT Fusion} and BGE-L. Bold indicates the best performance within each backbone pair. Abbreviations: MSLS-Blur (MSLS-B), MSLS-Weather (MSLS-W), Amstertime-La (Ams.-La), MSLS Challenge (MSLS Ch.).}
\label{tab:lavpr_cat_multimodal}
\renewcommand{\arraystretch}{1.1}
\setlength{\tabcolsep}{1.1pt}
\begin{tabular}{lr ccc ccc ccc ccc ccc}
\toprule
\textbf{Method} & \textbf{Dim} & \multicolumn{3}{c}{\textbf{MSLS-B}} & \multicolumn{3}{c}{\textbf{MSLS-W}} & \multicolumn{3}{c}{\textbf{Amst.-La}} & \multicolumn{3}{c}{\textbf{MSLS Ch.}} & \multicolumn{3}{c}{\textbf{Pitts30}} \\
\cmidrule(lr){3-5} \cmidrule(lr){6-8} \cmidrule(lr){9-11} \cmidrule(lr){12-14} \cmidrule(lr){15-17}
& ($d$) & R@1 & R@5 & R@10 & R@1 & R@5 & R@10 & R@1 & R@5 & R@10 & R@1 & R@5 & R@10 & R@1 & R@5 & R@10 \\
\midrule
NetVLAD & 32768 & 24.4 & 34.9 & 39.5 & 25.6 & 31.4 & 38.4 & 19.4 & 48.4 & 51.6 & 27.9 & 37.0 & 41.8 & \textbf{82.6} & 91.6 & 94.3 \\
\rowcolor{highlight} La-NetVLAD & +1024 & \textbf{64.0} & \textbf{75.6} & \textbf{82.6} & \textbf{60.5} & \textbf{76.7} & \textbf{82.6} & \textbf{58.1} & \textbf{80.6} & \textbf{80.6} & \textbf{33.0} & \textbf{46.0} & \textbf{51.2} & 81.8 & \textbf{92.2} & \textbf{94.5} \\
\midrule
CosPlace & 512 & 46.5 & 66.3 & 72.1 & 67.4 & 84.9 & 87.2 & 48.4 & 67.7 & 74.2 & 57.4 & 69.9 & 73.9 & 89.0 & 94.7 & 96.0 \\
\rowcolor{highlight} La-CosPlace & +1024 & \textbf{67.4} & \textbf{81.4} & \textbf{87.}2 & \textbf{87.2} & \textbf{96.5} & \textbf{96.5} & \textbf{54.8} & \textbf{80.6} & \textbf{87.1} & \textbf{60.7} & \textbf{73.7} & \textbf{77.3} & \textbf{89.3} & \textbf{95.0} & \textbf{96.2} \\
\midrule
EigenPlaces & 2048 & 51.2 & 76.7 & 80.2 & 75.6 & 88.4 & 91.9 & 41.9 & 64.5 & 83.9 & 62.1 & 73.9 & 76.4 & 90.6 & 95.4 & 96.8 \\
\rowcolor{highlight} La-EigenPlace & +1024 & \textbf{73.3} & \textbf{87.2} & \textbf{88.4} & \textbf{87.2} & \textbf{94.2} & \textbf{95.3} & \textbf{48.4} & \textbf{87.1} & \textbf{90.3} & \textbf{64.2} & \textbf{75.5} & \textbf{79.1} & \textbf{90.8} & \textbf{95.7} & 96.8 \\
\midrule
MixVPR & 512 & 62.8 & 76.7 & 82.6 & 70.9 & 86.0 & 90.7 & 41.9 & 80.6 & 83.9 & 59.4 & 72.3 & 77.3 & \textbf{90.6} &\textbf{95.6} & 96.3 \\
\rowcolor{highlight} La-MixVPR & +1024 & \textbf{83.7} & \textbf{94.2} & \textbf{96.5} & \textbf{83.7} & \textbf{94.2} & \textbf{96.5} & \textbf{58.1} & \textbf{90.3} & \textbf{93.5} & \textbf{61.5} & \textbf{74.9} & \textbf{79.9} & 89.9 & 95.3 & \textbf{96.4} \\
\midrule
MixVPR & 4096 & 70.9 & 82.6 & 87.2 & 77.9 & 83.7 & 88.4 & 54.8 & 80.6 & 83.9 & 63.4 & 76.0 & 79.2 & \textbf{91.6} & \textbf{95.6} & 96.4 \\
\rowcolor{highlight} La-MixVPR & +1024 & \textbf{86.0} & \textbf{98.8} & \textbf{98.8} & \textbf{89.5} & \textbf{95.3} & \textbf{97.7} & \textbf{58.1} & \textbf{87.1} & \textbf{93.5} & \textbf{64.0} & \textbf{76.3} & \textbf{81.1} & 90.8 & 95.2 & 96.4 \\
\midrule
SALAD & 8448 & 91.9 & 98.8 & 98.8 & 90.7 & 96.5 & 97.7 & 58.1 & 90.3 & 93.5 & \textbf{74.6} & \textbf{88.4} & \textbf{91.0} & \textbf{92.2} & \textbf{96.1} & \textbf{97.4} \\
\rowcolor{highlight} La-SALAD & +1024 & \textbf{93.0} & 98.8 & \textbf{100} & \textbf{94.2} & \textbf{98.8} & \textbf{100} & \textbf{61.3} & \textbf{93.5} & 93.5 & 73.9 & 87.7 & 90.1 & 91.5 & 96.1 & 97.1 \\
\midrule
CricaVPR & 10752 & 91.9 & \textbf{98.8} & 98.8 & 88.4 & 94.2 & 97.7 & \textbf{64.5} & 90.3 & 93.5 & \textbf{69.8} & \textbf{82.2} & 85.4 & \textbf{93.2} & \textbf{96.7} & \textbf{97.7} \\
\rowcolor{highlight} La-CricaVPR & +1024 & 91.9 & 97.7 & 98.8 & \textbf{93.0} & \textbf{98.8} & \textbf{100} & 58.1 & \textbf{93.5} & 93.5 & 67.2 & 81.7 & \textbf{85.8} & 91.8 & 96.3 & 97.5 \\
\bottomrule
\end{tabular}
\end{table*}
\begin{table*}[h!]
\centering
\scriptsize
\caption{Impact of Textual Encoders on Visual Place Recognition.}\label{tab:effect_of_text_encoder}
\renewcommand{\arraystretch}{1.1}
\setlength{\tabcolsep}{2pt}
\begin{tabular}{ll c ccc ccc ccc}
\toprule
\textbf{Fusion} & \textbf{Text Encoder} & \textbf{Dim} & \multicolumn{2}{c}{\textbf{MSLS-Blur}} & \multicolumn{2}{c}{\textbf{MSLS-Weather}} & \multicolumn{2}{c}{\textbf{Amstertime-La}} & \multicolumn{2}{c}{\textbf{Pitts30}} \\
& & & R@1 & R@5 & R@1 & R@5 & R@1 & R@5 & R@1 & R@5 \\
\midrule
\rowcolor{gray!10} \multicolumn{10}{l}{\textit{Text-only Baselines}} \\
None & BGE-S & 384 & 57.0 & 80.2 & 57.0 & 80.2 & 25.8 & 58.1 & 37.3 & 60.9 \\
None & BGE-L & 1024 & 54.7 & 74.4 & 54.7 & 74.4 & 38.7 & 67.7 & 36.5 & 62.4 \\
None & ModernBert-L & 1024 & 60.5 & 76.7 & 60.5 & 76.7 & 35.5 & 58.1 & 37.1 & 61.8 \\
\midrule
\rowcolor{gray!10} \multicolumn{10}{l}{\textit{Vision-only Baseline: MixVPR}} \\
None & -- & 512 & 62.8 & 76.7 & 70.9 & 86.0 & 41.9 & 80.6 & \textbf{90.6} & \textbf{95.6} \\
\midrule
\rowcolor{gray!10} \multicolumn{10}{l}{\textit{Multi-Modal (MixVPR-512 with Text Encoders)}} \\
\multirow{3}{*}{\textbf{CAT}} 
& BGE-S & +384 & 83.7 & 94.2 & 88.4 & \textbf{96.5} & 58.1 & 87.1 & 90.1 & 95.2 \\
& BGE-L & +1024 & 83.7 & 94.2 & 83.7 & 94.2 & 58.1 & \textbf{90.3} & 89.9 & 95.3 \\
& ModernBert-L & +1024 & 81.4 & 91.9 & 86.0 & 95.3 & \textbf{61.3} & 83.9 & 90.7 & 95.2 \\
\midrule
\multirow{3}{*}{\textbf{ADS+LLP}} 
& BGE-S & +384 & 87.2 & 95.3 & \textbf{89.5} & \textbf{96.5} & 48.4 & 83.9 & 90.0 & 95.1 \\
& BGE-L & +1024 & \textbf{89.5} & 96.5 & 88.4 & 94.2 & 58.1 & \textbf{90.3} & 88.8 & 94.8 \\
& ModernBert-L & +1024 & 87.2 & \textbf{98.8} & 88.4 & \textbf{96.5} & \textbf{61.3} & \textbf{90.3} & 88.6 & 94.7 \\
\bottomrule
\end{tabular}
\end{table*}

\section{Efficiency-Performance Trade-off Analysis}\label{sec:appendix_complexity}
Table 7 in the main manuscript presents a comparative analysis between the baseline CricaVPR variants and our proposed Language-Augmented (LA) Small variant utilizing CAT fusion. For this evaluation, we adopt the model parameter counts for CricaVPR as reported in \cite{lu2024cricavpr}. For CricaVPR-Small which uses DinoV2-Small as its backbone \cite{oquab2023dinov2, caron2021emerging}, similar to CricaVPR, we added the adapters to its parameters count. BGE-Small parameter count is based on ~\cite{bge_embedding}, where we used the model's english version. 

Following the methodology of \cite{kaplan2020scaling}, the computational cost $C$ for a Transformer forward pass over a sequence of length $S$ is approximated by:
\begin{equation}
C \approx 2NS \text{ FLOPs}
\end{equation}
where $N$ denotes the number of non-embedding parameters, $S$ represents the sequence length in tokens, and the factor of 2 accounts for multiply-accumulate (MAC) operations, comprising one multiplication and one addition per token.

The CricaVPR-Small architecture employs a hidden dimension of 384 with 6 attention heads, whereas the CricaVPR-Base variant incorporates 768 dimensions and 12 attention heads. This architectural scaling yields a $4\times$ increase in floating-point operations (FLOPs), thereby providing enhanced representational capacity at the expense of substantially higher computational overhead. CricaVPR's input image resolution is 224x224.

For the BGE-Small model, the computational cost is estimated as $C \approx 2NS \approx 2.1$ GFLOPs, where $N \approx 21.2$M denotes the non-embedding parameters and $S = 50$ represents the sequence length (average caption length as seen in Table 3 in the main manuscript).

We evaluate the total computational complexity by partitioning the pipeline into two primary stages: 
\begin{itemize}
 \item {Inference:} The forward pass through the CricaVPR backbone to extract the image embedding vector.
    \item {Retrieval:} The vector search phase, which calculates similarity metrics (e.g., cosine similarity) between the query embedding and the database entries.
\end{itemize}
The total parameter count for our \textit{LA-Crica-Small} model is calculated as the summation of the parameters from the CricaVPR-Small and BGE-Small architectures.
\section{Multi-Modal Place Recognition: Implementation Details}\label{subsec:appendix_multimodal_impl}
We trained several late fusion architectures using frozen VPR backbones and a frozen text encoder backbone (BGE). The fusion architectures evaluated include CAT, PA, MLP, and ADS. All techniques were trained on the same dataset, adhering to the standard GSV-Cities framework augmented with our textual descriptions, which provides a highly accurate dataset comprising 67k places represented by 529k images. We employed the Multi-Similarity loss function for training, using $\alpha$=1, $\beta$=50 and margin=0. Each training batch contained $P = 120$ places, with 4 images per place, yielding mini-batches of 480 images. Optimization was performed using Stochastic Gradient Descent (SGD) with a momentum of 0.9 and weight decay of 0.001. The initial learning rate was set to 0.05 and reduced by a factor of 3 every 2 epochs. Training proceeded for a maximum of 10 epochs on an NVIDIA RTX 3090 GPU equipped with 24GB VRAM.

\begin{figure}[b]
    \centering
    \includegraphics[scale=0.5]{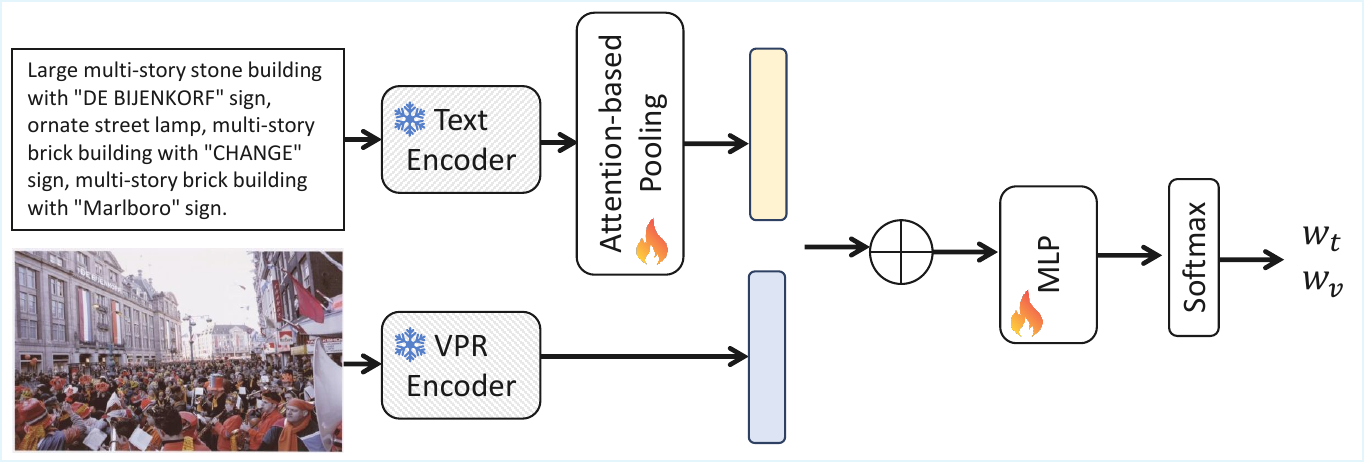}
    \caption{Illustration of the ADS+LLP Fusion Method.}
    \label{fig:llp_ads_arch}
\end{figure}

Figure~\ref{fig:llp_ads_arch} depicts the ADS+LLP fusion approach. The LLP employs an attention mechanism in which each token is processed through a linear layer, after which a softmax function calculates the importance score for each token. The embedding of each token (excluding the CLS token) is then weighted by its corresponding importance value to generate a pooled representation. This pooled token is concatenated with the CLS token and subsequently passed through an additional linear layer followed by a Tanh activation function, yielding the final embedding.
\section{Cross-Modal Place Recognition: Implementation Details}\label{subsec:appendix_crossmodal_impl}
We trained multiple architectures (learned pooling, LoRA, full fine-tuning) with different loss functions (Contrastive and Multi-Similarity loss) across various VLM encoders, using $\alpha$=2, $\beta$=40 and margin=0.5. In all cases, training was applied to both the visual and textual encoders. Training batches consisted of $P = 60$ places, each represented by 4 images, resulting in mini-batches of 240 images. We employed Stochastic Gradient Descent (SGD) with momentum 0.9 and weight decay of 0.001. The initial learning rate of 0.05 was divided by 3 every 2 epochs, and training was conducted for a maximum of 10 epochs. Experiments were performed on an NVIDIA H200 GPU with 141GB VRAM. For all models used for evaluation (Blip, SigLip etc.) we have used the Base version.

We further investigated an approach involving two separate pooling layers trained on top of a frozen image encoder (MixVPR, DinoV2) and a frozen text encoder (BGE). We observed that the training process did not converge and yielded poor performance. 
\section{Qualitative Results for Multi-Modal and Cross-Modal Place Recognition}
\label{subsec:appendix_qualitative_results}
\begin{figure}
\centering
\begin{subfigure}{1\textwidth}
  \centering
  \includegraphics[width=1\linewidth]{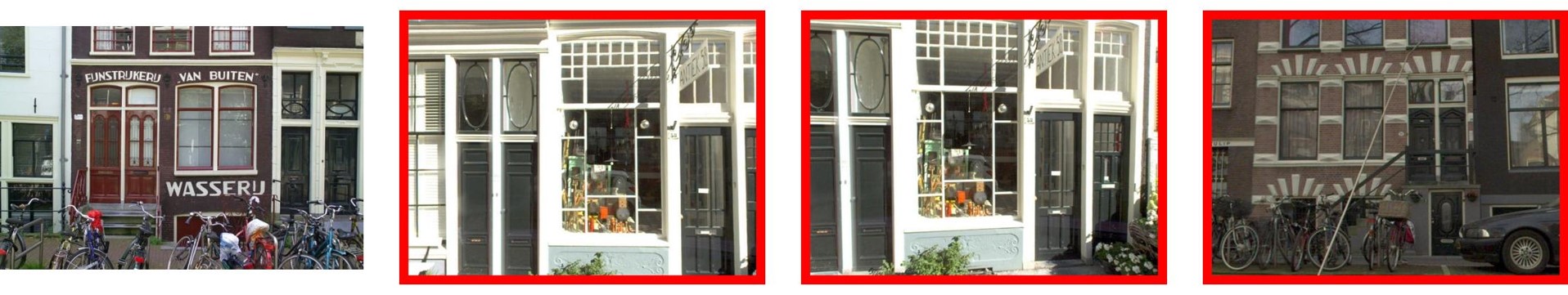}
  \caption{MixVPR}
  \label{fig:amstertime_mix1}
\end{subfigure}


\begin{subfigure}{1\textwidth}
  \centering
  \includegraphics[width=1\linewidth]{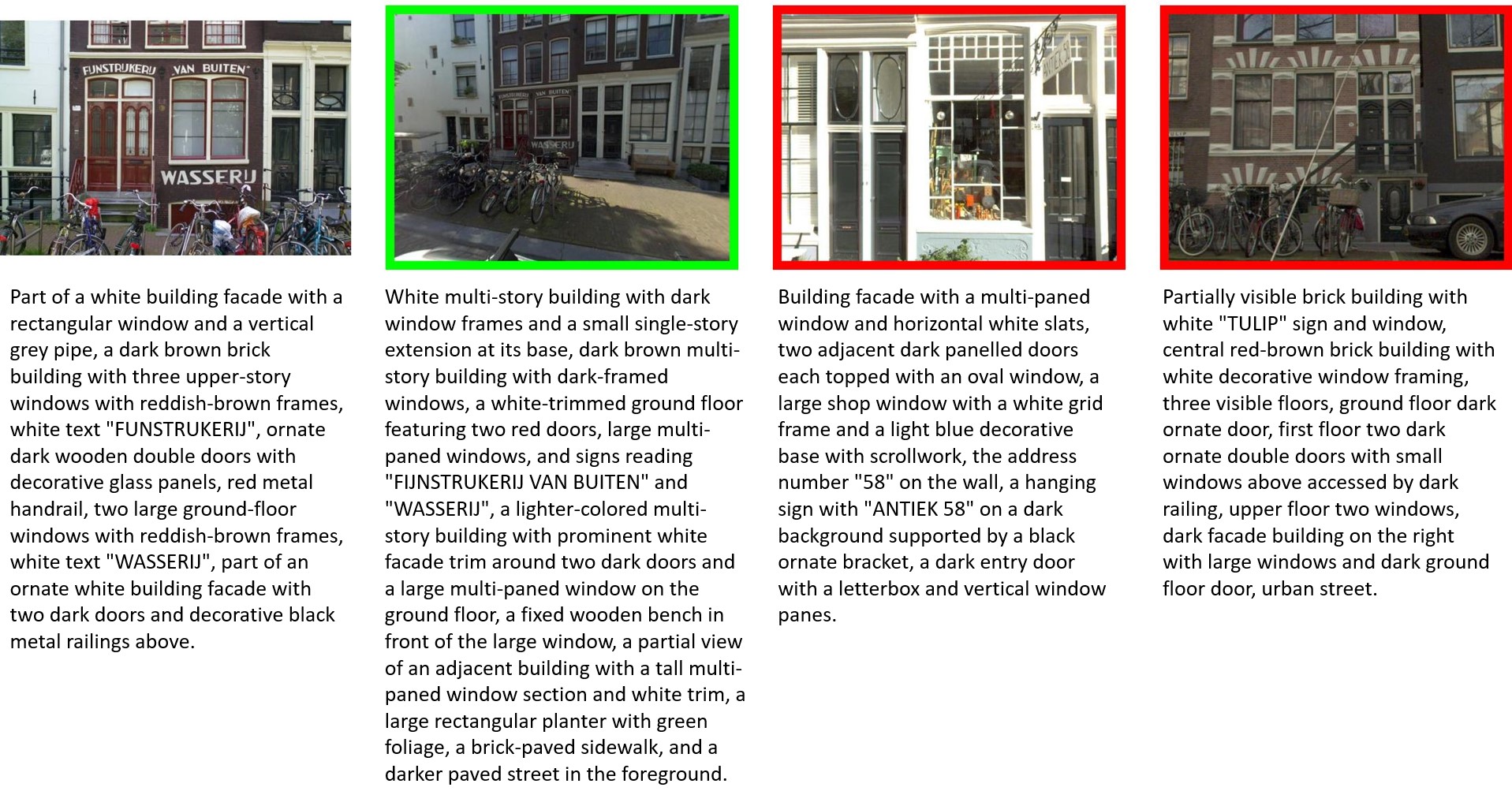}
  \caption{La-MixVPR}
  \label{fig:sub2}
\end{subfigure}
\caption{Qualitative comparison between MixVPR (top) and La-MixVPR (bottom) on the Amstertime dataset. La-MixVPR uses BGE-Large as the text encoder with CAT fusion. From left to right: Query, Top-1, Top-2, and Top-3 retrieved images. Green borders indicate a correct match, while red borders indicate an incorrect match.}
\label{fig:qex1}
\end{figure}

\begin{figure}
\centering
\begin{subfigure}{1\textwidth}
  \centering
  \includegraphics[width=1\linewidth]{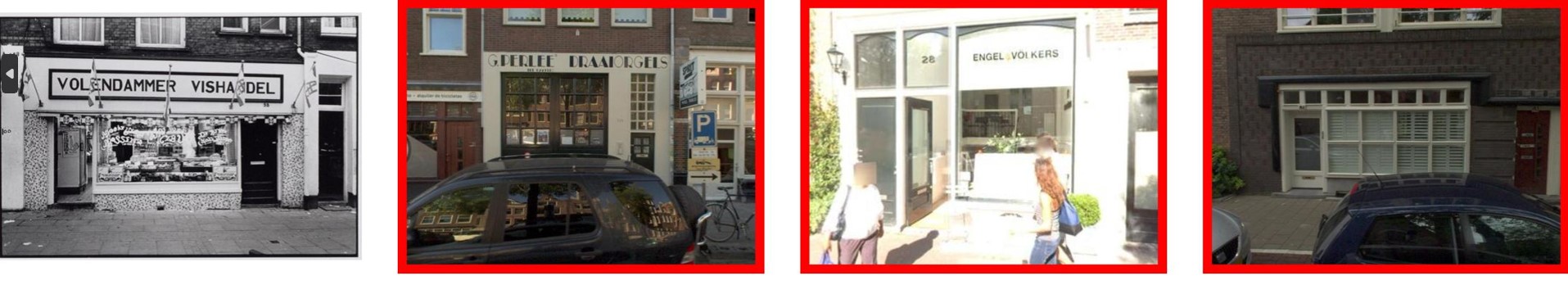}
  \caption{MixVPR}
  \label{fig:amstertime_mix1}
\end{subfigure}


\begin{subfigure}{1\textwidth}
  \centering
  \includegraphics[width=1\linewidth]{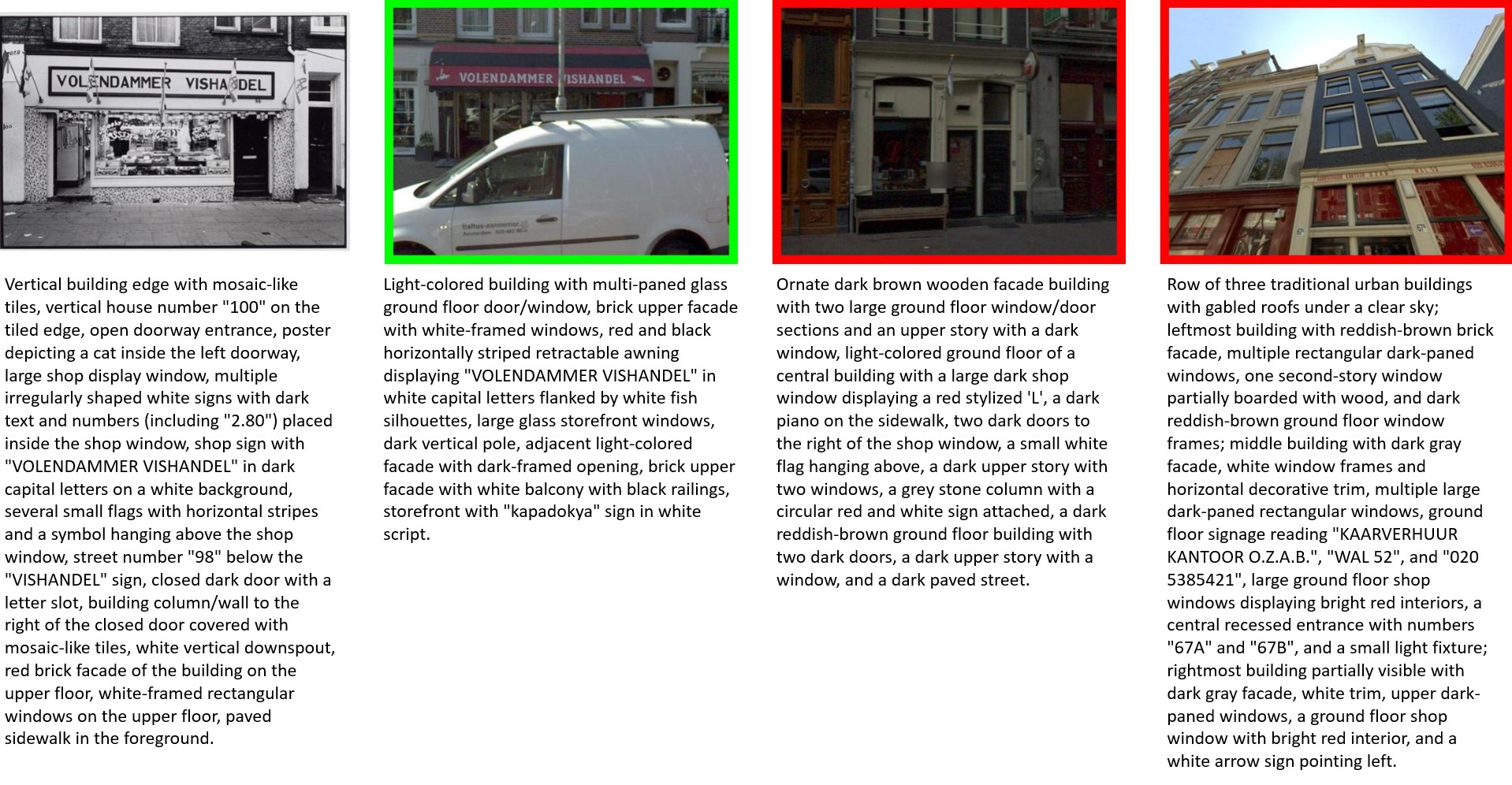}
  \caption{La-MixVPR}
  \label{fig:sub2}
\end{subfigure}
\caption{Qualitative comparison between MixVPR (top) and La-MixVPR (bottom) on the Amstertime dataset. La-MixVPR uses BGE-Large as the text encoder with CAT fusion. From left to right: Query, Top-1, Top-2, and Top-3 retrieved images. Green borders indicate a correct match, while red borders indicate an incorrect match.}
\label{fig:qex2}
\end{figure}

\begin{figure}
\centering
\begin{subfigure}{1\textwidth}
  \centering
  \includegraphics[width=1\linewidth]{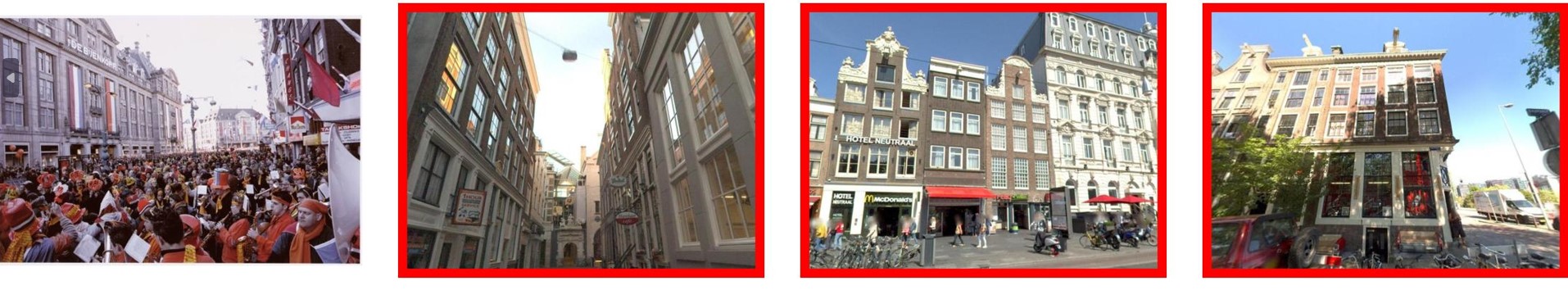}
  \caption{MixVPR}
  \label{fig:amstertime_mix1}
\end{subfigure}


\begin{subfigure}{1\textwidth}
  \centering
  \includegraphics[width=1\linewidth]{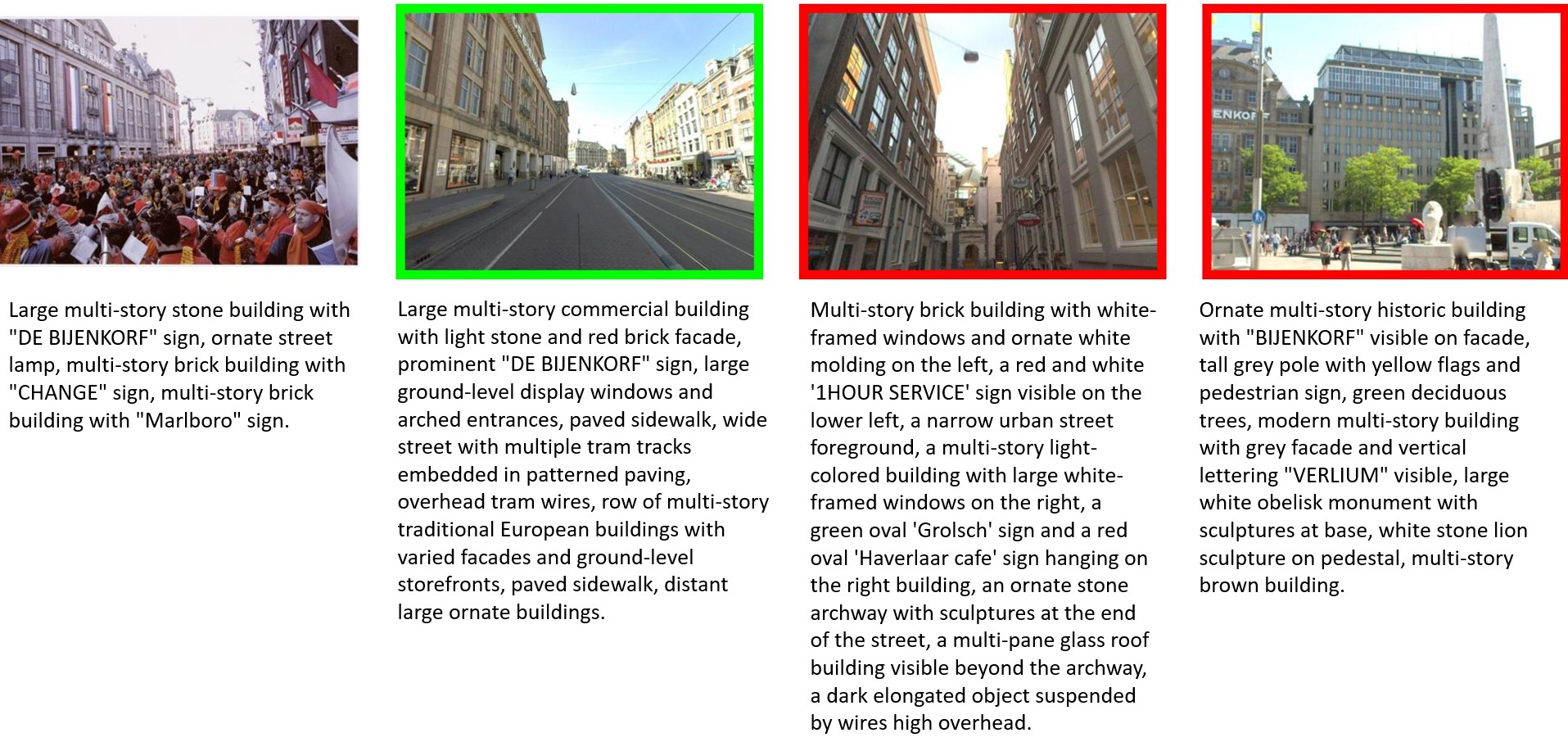}
  \caption{La-MixVPR}
  \label{fig:sub2}
\end{subfigure}
\caption{Qualitative comparison between MixVPR (top) and La-MixVPR (bottom) on the Amstertime dataset. La-MixVPR uses BGE-Large as the text encoder with CAT fusion. From left to right: Query, Top-1, Top-2, and Top-3 retrieved images. Green borders indicate a correct match, while red borders indicate an incorrect match.}
\label{fig:qex3}
\end{figure}

\begin{figure}[h]
    \centering
    \includegraphics[scale=0.26]{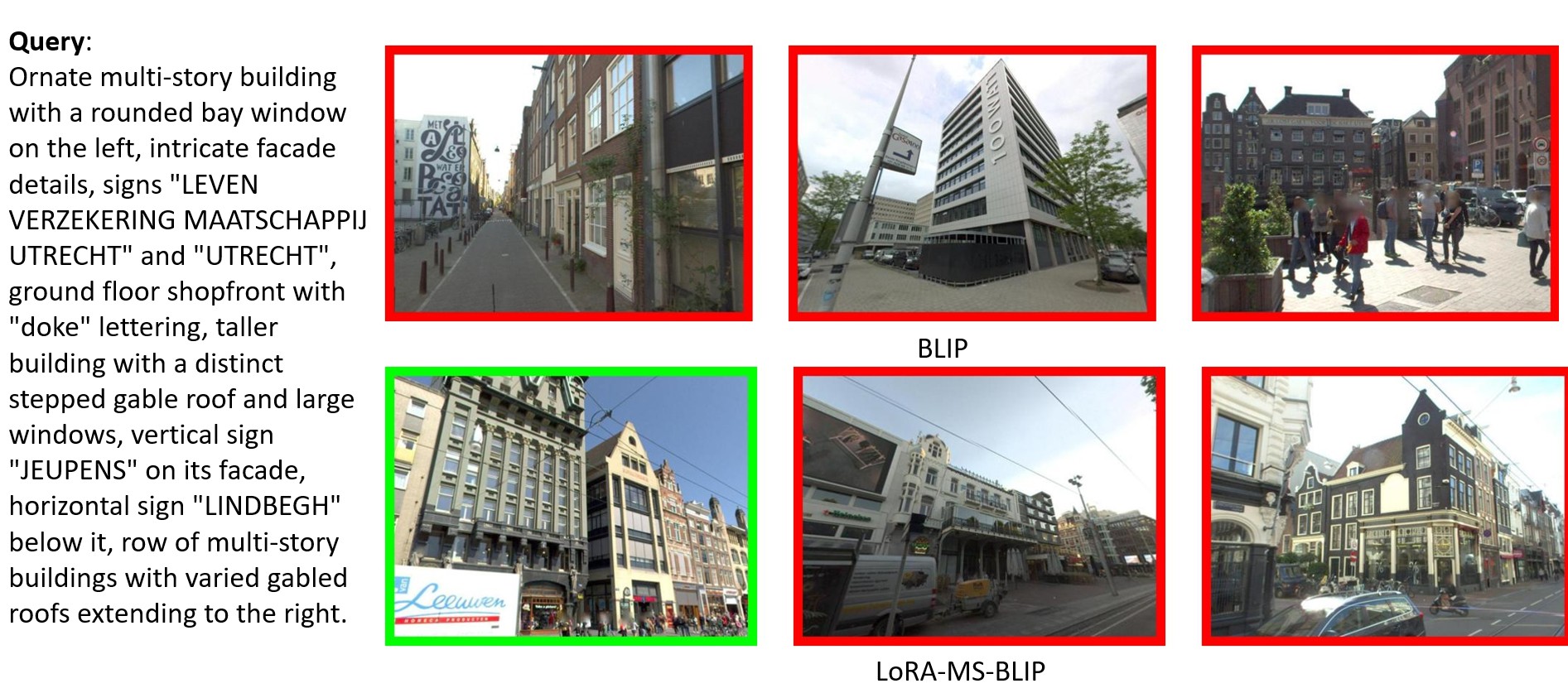}
    \caption{Cross-Modal retrieval (Amstertime dataset): we use a textual query (left) to retrieve geo-tagged images from a reference database. We present the top-3 results when using BLIP (top) and LoRA-MS-BLIP (bottom). Green borders indicate a correct match, while red borders indicate an incorrect match.}
    \label{fig:cross_modal_ex1}
\end{figure}

\begin{figure}[h]
    \centering
    \includegraphics[scale=0.26]{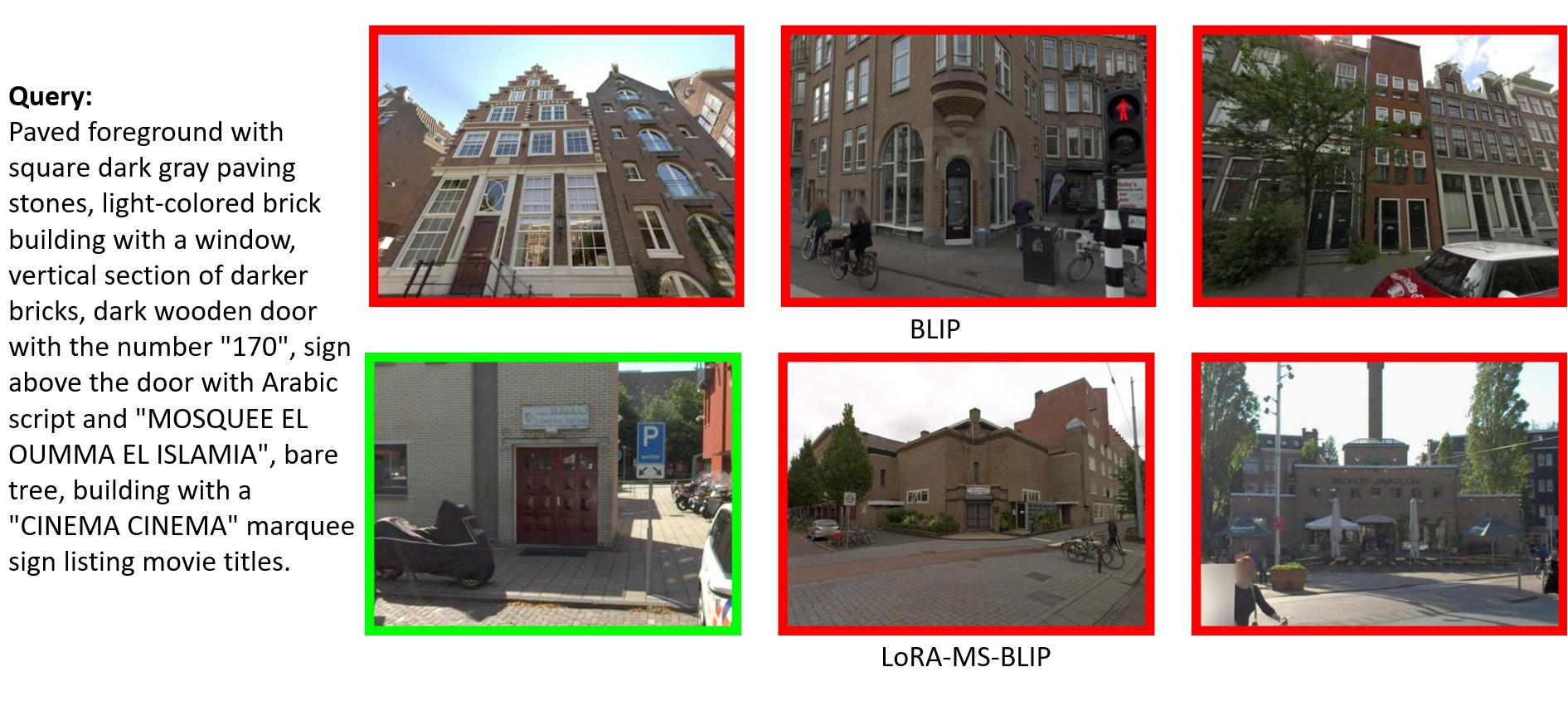}
    \caption{Cross-Modal retrieval (Amstertime dataset): we use a textual query (left) to retrieve geo-tagged images from a reference database. We present the top-3 results when using BLIP (top) and LoRA-MS-BLIP (bottom). Green borders indicate a correct match, while red borders indicate an incorrect match.}
    \label{fig:cross_modal_ex2}
\end{figure}

\begin{figure}[h]
    \centering
    \includegraphics[scale=0.26]{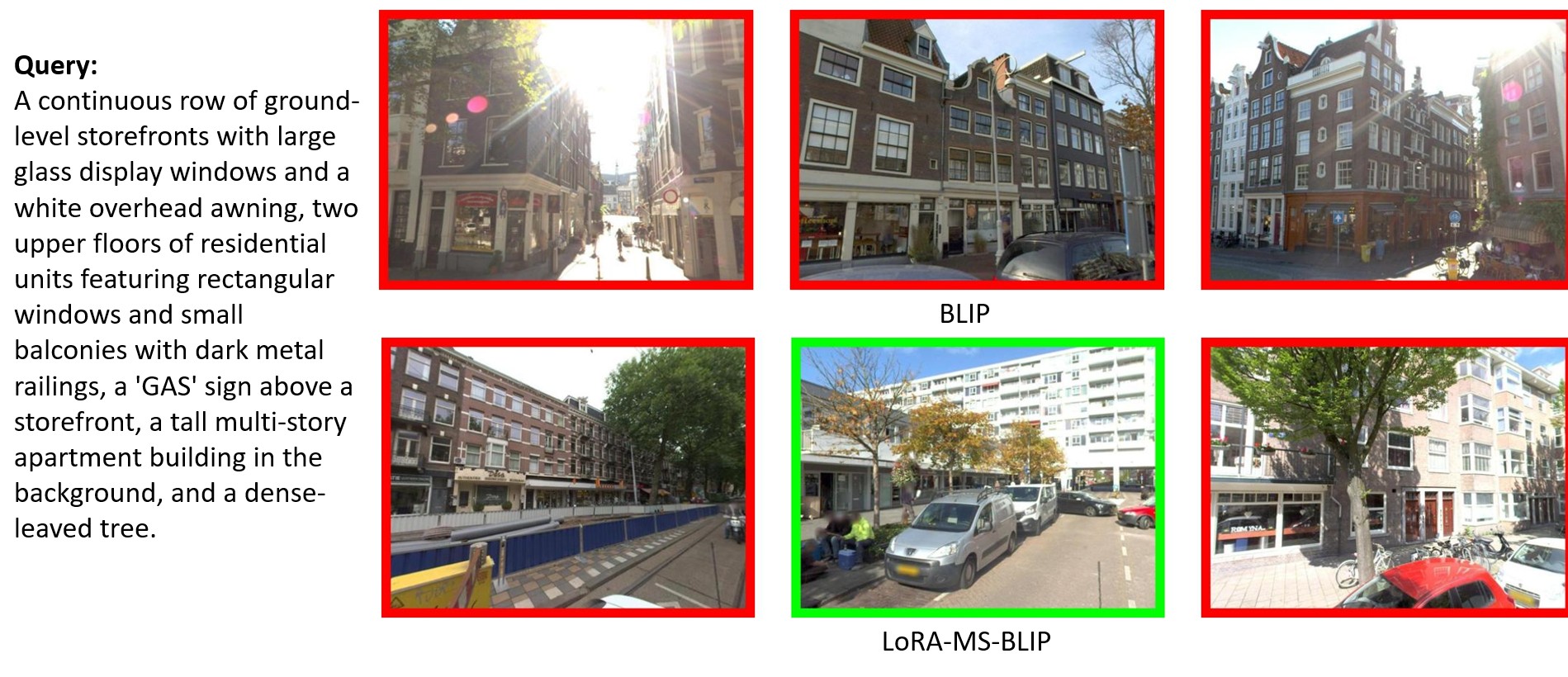}
    \caption{Cross-Modal retrieval (Amstertime dataset): we use a textual query (left) to retrieve geo-tagged images from a reference database. We present the top-3 results when using BLIP (top) and LoRA-MS-BLIP (bottom). Green borders indicate a correct match, while red borders indicate an incorrect match.}
    \label{fig:cross_modal_ex3}
\end{figure}

Figures~\ref{fig:qex1}-\ref{fig:qex3} show selected examples of retrieval results using MixVPR and La-MixVPR (ours) for multi-modal place recognition. Figures~\ref{fig:cross_modal_ex1}-\ref{fig:cross_modal_ex3} compare cross-modal place recognition with BLIP and LoRA-MS-BLIP.


\clearpage
\bibliographystyle{splncs04}


\begin{thebibliography}{10}
\providecommand{\url}[1]{\texttt{#1}}
\providecommand{\urlprefix}{URL }
\providecommand{\doi}[1]{https://doi.org/#1}

\bibitem{abdin2024phi-3}
Abdin, M.I., Ade~Jacobs, S., Awan, A.A., et~al.: Phi-3 technical report: A highly capable language model locally on your phone. Tech. Rep. MSR-TR-2024-12, Microsoft (August 2024), \url{https://arxiv.org/abs/2404.14219}

\bibitem{Alibey_2022_gsvcities}
Ali-bey, A., Chaib-draa, B., Giguere, P.: Gsv-cities: Toward appropriate supervised visual place recognition. Neurocomputing  (2022)

\bibitem{ali2023mixvpr}
Ali-Bey, A., Chaib-Draa, B., Giguere, P.: Mixvpr: Feature mixing for visual place recognition. In: Proceedings of the IEEE/CVF winter conference on applications of computer vision. pp. 2998--3007 (2023)

\bibitem{arandjelovic2016netvlad}
Arandjelovic, R., Gronat, P., Torii, A., Pajdla, T., Sivic, J.: Netvlad: Cnn architecture for weakly supervised place recognition. In: Proceedings of the IEEE conference on computer vision and pattern recognition. pp. 5297--5307 (2016)

\bibitem{berton2022rethinking}
Berton, G., Masone, C., Caputo, B.: Rethinking visual geo-localization for large-scale applications. In: Proceedings of the IEEE/CVF Conference on Computer Vision and Pattern Recognition. pp. 4878--4888 (2022)

\bibitem{berton2023eigenplaces}
Berton, G., Trivigno, G., Caputo, B., Masone, C.: Eigenplaces: Training viewpoint robust models for visual place recognition. In: Proceedings of the IEEE/CVF International Conference on Computer Vision. pp. 11080--11090 (2023)

\bibitem{carion2025sam}
Carion, N., Gustafson, L., Hu, Y.T., Debnath, S., Hu, R., Suris, D., Ryali, C., Alwala, K.V., Khedr, H., Huang, A., et~al.: Sam 3: Segment anything with concepts. arXiv preprint arXiv:2511.16719  (2025)

\bibitem{ncarlevaris-2015a}
Carlevaris-Bianco, N., Ushani, A.K., Eustice, R.M.: University of {Michigan} {North} {Campus} long-term vision and lidar dataset. International Journal of Robotics Research  \textbf{35}(9),  1023--1035 (2015)

\bibitem{caron2021emerging}
Caron, M., Touvron, H., Misra, I., J{\'e}gou, H., Mairal, J., Bojanowski, P., Joulin, A.: Emerging properties in self-supervised vision transformers. In: Proceedings of the IEEE/CVF international conference on computer vision. pp. 9650--9660 (2021)

\bibitem{eva02}
Fang, Y., Sun, Q., Wang, X., Huang, T., Wang, X., Cao, Y.: Eva-02: A visual representation for neon genesis. Image and Vision Computing p. 105171 (2024)

\bibitem{google2025gemini}
Google: Gemini 2.5 flash (2025), \url{https://deepmind.google/technologies/gemini/}, large Language Model

\bibitem{hausler2021patch}
Hausler, S., Garg, S., Xu, M., Milford, M., Fischer, T.: Patch-netvlad: Multi-scale fusion of locally-global descriptors for place recognition. In: Proceedings of the IEEE/CVF conference on computer vision and pattern recognition. pp. 14141--14152 (2021)

\bibitem{hoffer2015deep}
Hoffer, E., Ailon, N.: Deep metric learning using triplet network. In: International workshop on similarity-based pattern recognition. pp. 84--92. Springer (2015)

\bibitem{hong2019textplace}
Hong, Z., Petillot, Y., Lane, D., Miao, Y., Wang, S.: Textplace: Visual place recognition and topological localization through reading scene texts. In: Proceedings of the IEEE/CVF International Conference on Computer Vision. pp. 2861--2870 (2019)

\bibitem{hu2022lora}
Hu, E.J., Shen, Y., Wallis, P., Allen-Zhu, Z., Li, Y., Wang, S., Wang, L., Chen, W., et~al.: Lora: Low-rank adaptation of large language models. ICLR  \textbf{1}(2), ~3 (2022)

\bibitem{hu2020dasgil}
Hu, H., Qiao, Z., Cheng, M., Liu, Z., Wang, H.: Dasgil: Domain adaptation for semantic and geometric-aware image-based localization. IEEE Transactions on Image Processing  \textbf{30},  1342--1353 (2020)

\bibitem{huang2024dino}
Huang, G., Zhou, Y., Hu, X., Zhang, C., Zhao, L., Gan, W.: Dino-mix enhancing visual place recognition with foundational vision model and feature mixing. Scientific Reports  \textbf{14}(1),  22100 (2024)

\bibitem{huang2025survey}
Huang, L., Yu, W., Ma, W., Zhong, W., Feng, Z., Wang, H., Chen, Q., Peng, W., Feng, X., Qin, B., et~al.: A survey on hallucination in large language models: Principles, taxonomy, challenges, and open questions. ACM Transactions on Information Systems  \textbf{43}(2),  1--55 (2025)

\bibitem{izquierdo2024optimal}
Izquierdo, S., Civera, J.: Optimal transport aggregation for visual place recognition. In: Proceedings of the ieee/cvf conference on computer vision and pattern recognition. pp. 17658--17668 (2024)

\bibitem{kaplan2020scaling}
Kaplan, J., McCandlish, S., Henighan, T., Brown, T.B., Chess, B., Child, R., Gray, S., Radford, A., Wu, J., Amodei, D.: Scaling laws for neural language models. arXiv preprint arXiv:2001.08361  (2020)

\bibitem{kolmet2022text2pos}
Kolmet, M., Zhou, Q., O{\v{s}}ep, A., Leal-Taix{\'e}, L.: Text2pos: Text-to-point-cloud cross-modal localization. In: Proceedings of the IEEE/CVF Conference on Computer Vision and Pattern Recognition. pp. 6687--6696 (2022)

\bibitem{komorowski2021minkloc++}
Komorowski, J., Wysocza{\'n}ska, M., Trzcinski, T.: Minkloc++: lidar and monocular image fusion for place recognition. In: 2021 International Joint Conference on Neural Networks (IJCNN). pp.~1--8. IEEE (2021)

\bibitem{lai2022adafusion}
Lai, H., Yin, P., Scherer, S.: Adafusion: Visual-lidar fusion with adaptive weights for place recognition. IEEE Robotics and Automation Letters  \textbf{7}(4),  12038--12045 (2022)

\bibitem{li2022blip}
Li, J., Li, D., Xiong, C., Hoi, S.: Blip: Bootstrapping language-image pre-training for unified vision-language understanding and generation. In: International conference on machine learning. pp. 12888--12900. PMLR (2022)

\bibitem{li2025unified}
Li, K., Ou, Y., Ning, J., Kong, F., Cai, H., Li, H.: Unified depth-guided feature fusion and reranking for hierarchical place recognition. Sensors  \textbf{25}(13), ~4056 (2025)

\bibitem{li2025place}
Li, Z., Shang, T., Xu, P., Deng, Z.: Place recognition meet multiple modalities: a comprehensive review, current challenges and future development. Artificial Intelligence Review  \textbf{58}(11), ~363 (2025)

\bibitem{liu2025text}
Liu, D., Huang, S., Li, W., Shen, S., Wang, C.: Text to point cloud localization with multi-level negative contrastive learning. In: Proceedings of the AAAI Conference on Artificial Intelligence. vol.~39, pp. 5397--5405 (2025)

\bibitem{lu2024cricavpr}
Lu, F., Lan, X., Zhang, L., Jiang, D., Wang, Y., Yuan, C.: Cricavpr: Cross-image correlation-aware representation learning for visual place recognition. In: Proceedings of the IEEE/CVF Conference on Computer Vision and Pattern Recognition. pp. 16772--16782 (2024)

\bibitem{lu2024towards}
Lu, F., Zhang, L., Lan, X., Dong, S., Wang, Y., Yuan, C.: Towards seamless adaptation of pre-trained models for visual place recognition. In: The Twelfth International Conference on Learning Representations (2024), \url{https://openreview.net/forum?id=TVg6hlfsKa}

\bibitem{maddern20171}
Maddern, W., Pascoe, G., Linegar, C., Newman, P.: 1 year, 1000 km: The oxford robotcar dataset. The International Journal of Robotics Research  \textbf{36}(1),  3--15 (2017)

\bibitem{melekhin2025mssplace}
Melekhin, A., Yudin, D., Petryashin, I., Bezuglyj, V.: Mssplace: Multi-sensor place recognition with visual and text semantics. IEEE Access  (2025)

\bibitem{oquab2023dinov2}
Oquab, M., Darcet, T., Moutakanni, T., Vo, H., Szafraniec, M., Khalidov, V., Fernandez, P., Haziza, D., Massa, F., El-Nouby, A., et~al.: Dinov2: Learning robust visual features without supervision. arXiv preprint arXiv:2304.07193  (2023)

\bibitem{piasco2021improving}
Piasco, N., Sidib{\'e}, D., Gouet-Brunet, V., Demonceaux, C.: Improving image description with auxiliary modality for visual localization in challenging conditions. International Journal of Computer Vision  \textbf{129}(1),  185--202 (2021)

\bibitem{radenovic2018fine}
Radenovi{\'c}, F., Tolias, G., Chum, O.: Fine-tuning cnn image retrieval with no human annotation. IEEE transactions on pattern analysis and machine intelligence  \textbf{41}(7),  1655--1668 (2018)

\bibitem{clip}
Radford, A., Kim, J.W., Hallacy, C., Ramesh, A., Goh, G., Agarwal, S., Sastry, G., Askell, A., Mishkin, P., Clark, J., et~al.: Learning transferable visual models from natural language supervision. In: International conference on machine learning. pp. 8748--8763. PMLR (2021)

\bibitem{shang2025bridging}
Shang, T., Li, Z., Xu, P., Qiao, J., Chen, G., Ruan, Z., Hu, W.: Bridging text and vision: A multi-view text-vision registration approach for cross-modal place recognition. arXiv preprint arXiv:2502.14195  (2025)

\bibitem{torii201524}
Torii, A., Arandjelovic, R., Sivic, J., Okutomi, M., Pajdla, T.: 24/7 place recognition by view synthesis. In: Proceedings of the IEEE conference on computer vision and pattern recognition. pp. 1808--1817 (2015)

\bibitem{pitts}
Torii, A., Sivic, J., Pajdla, T., Okutomi, M.: Visual place recognition with repetitive structures. In: Proceedings of the IEEE conference on computer vision and pattern recognition. pp. 883--890 (2013)

\bibitem{tschannen2025siglip}
Tschannen, M., Gritsenko, A., Wang, X., Naeem, M.F., Alabdulmohsin, I., Parthasarathy, N., Evans, T., Beyer, L., Xia, Y., Mustafa, B., et~al.: Siglip 2: Multilingual vision-language encoders with improved semantic understanding. Localization, and Dense Features  \textbf{6} (2025)

\bibitem{wang2018cosface}
Wang, H., Wang, Y., Zhou, Z., Ji, X., Gong, D., Zhou, J., Li, Z., Liu, W.: Cosface: Large margin cosine loss for deep face recognition. In: Proceedings of the IEEE conference on computer vision and pattern recognition. pp. 5265--5274 (2018)

\bibitem{wang2024qwen2}
Wang, P., Bai, S., Tan, S., Wang, S., Fan, Z., Bai, J., Chen, K., Liu, X., Wang, J., Ge, W., et~al.: Qwen2-vl: Enhancing vision-language model's perception of the world at any resolution. arXiv preprint arXiv:2409.12191  (2024)

\bibitem{wang2019multi}
Wang, X., Han, X., Huang, W., Dong, D., Scott, M.R.: Multi-similarity loss with general pair weighting for deep metric learning. In: Proceedings of the IEEE/CVF conference on computer vision and pattern recognition. pp. 5022--5030 (2019)

\bibitem{msls}
Warburg, F., Hauberg, S., Lopez-Antequera, M., Gargallo, P., Kuang, Y., Civera, J.: Mapillary street-level sequences: A dataset for lifelong place recognition. In: Proceedings of the IEEE/CVF conference on computer vision and pattern recognition. pp. 2626--2635 (2020)

\bibitem{warner2025smarter}
Warner, B., Chaffin, A., Clavi{\'e}, B., Weller, O., Hallstr{\"o}m, O., Taghadouini, S., Gallagher, A., Biswas, R., Ladhak, F., Aarsen, T., et~al.: Smarter, better, faster, longer: A modern bidirectional encoder for fast, memory efficient, and long context finetuning and inference. In: Proceedings of the 63rd Annual Meeting of the Association for Computational Linguistics (Volume 1: Long Papers). pp. 2526--2547 (2025)

\bibitem{xia2024text2loc}
Xia, Y., Shi, L., Ding, Z., Henriques, J.F., Cremers, D.: Text2loc: 3d point cloud localization from natural language. In: Proceedings of the IEEE/CVF conference on computer vision and pattern recognition. pp. 14958--14967 (2024)

\bibitem{bge_embedding}
Xiao, S., Liu, Z., Zhang, P., Wang, N., , et~al.: C-pack: Packged resources for general chinese embedding. arXiv preprint arXiv:2309.07597  (2023)

\bibitem{xie2016semantic}
Xie, J., Kiefel, M., Sun, M.T., Geiger, A.: Semantic instance annotation of street scenes by 3d to 2d label transfer. In: Proceedings of the IEEE conference on Computer Vision and Pattern Recognition. pp. 3688--3697 (2016)

\bibitem{xu2025cmmloc}
Xu, Y., Qu, H., Liu, J., Zhang, W., Yang, X.: Cmmloc: Advancing text-to-pointcloud localization with cauchy-mixture-model based framework. In: Proceedings of the Computer Vision and Pattern Recognition Conference. pp. 6637--6647 (2025)

\bibitem{amstertime}
Yildiz, B., Khademi, S., Siebes, R.M., Van~Gemert, J.: Amstertime: A visual place recognition benchmark dataset for severe domain shift. In: 2022 26th International Conference on Pattern Recognition (ICPR). pp. 2749--2755. IEEE (2022)

\bibitem{zhou2023lcpr}
Zhou, Z., Xu, J., Xiong, G., Ma, J.: Lcpr: A multi-scale attention-based lidar-camera fusion network for place recognition. IEEE Robotics and Automation Letters  \textbf{9}(2),  1342--1349 (2023)

\bibitem{zhu2024minigpt}
Zhu, D., Chen, J., Shen, X., Li, X., Elhoseiny, M.: Mini{GPT}-4: Enhancing vision-language understanding with advanced large language models. In: The Twelfth International Conference on Learning Representations (2024), \url{https://openreview.net/forum?id=1tZbq88f27}

\end{thebibliography}

\setcounter{table}{0}
\end{document}